\documentclass{article}

\usepackage[preprint]{neurips_2020}
\usepackage[utf8]{inputenc} 
\usepackage[T1]{fontenc}    
\usepackage{hyperref}       
\usepackage{url}            
\usepackage{booktabs}       
\usepackage{amsfonts}       
\usepackage{nicefrac}       
\usepackage{microtype}      
\usepackage{graphicx}
\usepackage{subfigure}
\usepackage{booktabs} 
\usepackage{amsmath,amssymb}
\usepackage{amsthm}
\usepackage{booktabs}
\usepackage[table]{xcolor}   
\usepackage{xcolor}
\usepackage{slashbox,pict2e}
\usepackage{diagbox}
\usepackage{slashbox,multirow}
\usepackage{wrapfig}
\usepackage{enumitem}

\usepackage{hyperref}
\usepackage{bm,nicefrac}

\usepackage{amsfonts}

\usepackage{color}

\newtheorem{dfn}{Definition}

\newenvironment{itemize*}%
  {\begin{itemize}%
    \setlength{\itemsep}{1.5pt}%
    \setlength{\parskip}{2pt}}%
  {\end{itemize}}
\setcitestyle{numbers,open={[},close={]}}

\title{Physics-Guided Deep Learning for Dynamical Systems: A Survey}

\author{%
  Rui (Ray) Wang \\
  Computer Science and Engineering \\
  University of California, San Diego \\
  \texttt{ruw020@ucsd.edu} 
  \And
  Rose Yu \\
  Computer Science and Engineering \\
  University of California, San Diego \\
  \texttt{roseyu@ucsd.edu} 
}

\begin{document}
\maketitle
\begin{abstract} 
Modeling complex physical dynamics is a fundamental task in science and engineering. Traditional physics-based models are sample efficient, and interpretable but often rely on rigid assumptions. Furthermore,  direct numerical approximation is usually computationally intensive, requiring significant computational resources and expertise, and many real-world systems do not have fully-known governing laws.  While deep learning (DL) provides novel alternatives for efficiently recognizing complex patterns and emulating nonlinear dynamics, its predictions do not necessarily obey the governing laws of physical systems, nor do they generalize well across different systems. Thus, the study of physics-guided DL emerged and has gained great progress. Physics-guided DL aims to take the best from both physics-based modeling and state-of-the-art DL models to better solve scientific problems. In this paper, we provide a structured overview of existing methodologies of integrating prior physical knowledge or physics-based modeling into DL, with a special emphasis on learning dynamical systems. We also discuss the fundamental challenges and emerging opportunities in the area.
\end{abstract}

\section{Introduction}
Modeling complex physical dynamics over a wide range of spatial and temporal scales is a fundamental task in a wide range of fields including, for example, fluid dynamics \citep{strogatz2018nonlinear}, cosmology \citep{wainwright2005dynamical}, economics\citep{day1994complex}, and neuroscience \citep{izhikevich2007dynamical}.

Dynamical systems are mathematical objects that are used to describe the evolution of phenomena over time and space occurring in nature. Dynamical systems are commonly described with differential equations which are equations related to one or more unknown functions and their derivatives. 
\begin{dfn}\label{dfn:dynamics}
Fix an integer $k \geq 1$ and let $U$ denote an open subset of $\mathbb{R}^n$. Let $u: U \mapsto \mathbb{R}^m$ and we write $\bm{u} = (u^1, ..., u^m)$, where $x \in U$. Then an expression of the form 
\begin{equation}
    \mathcal{F}(D^k \bm{u}(x), D^{k-1} \bm{u}(x), ..., D\bm{u}(x), \bm{u}(x), x) = 0 \label{eqn:dynamics}
\end{equation}
is called a $k^{\text{-th}}$-order system of partial differential equation (or ordinary differential equation when $n=1$), where $\mathcal{F}: \mathbb{R}^{mn^k}\times \mathbb{R}^{mn^{k-1}}\times...\times\mathbb{R}^{mn}\times \mathbb{R}^{m}\times U \mapsto \mathbb{R}^m $.
\end{dfn}

$\mathcal{F}$ models the dynamics of a $n$-dimensional state $x \in \mathbb{R}^n$ and it can be either a linear or non-linear operator. Since most dynamics evolve over time, one of the variables of $u$ is usually the time dimension. In general, one must specify appropriate boundary and initial conditions of Equ.\ref{eqn:dynamics} to ensure the existence of a solution. Learning dynamical systems is to search for a model $\mathcal{F}$ that can accurately describe the behavior of the physical process insofar as we are interested. 

Physics as a discipline has a long tradition of using first principles to describe spatiotemporal dynamics. The laws of physics have greatly improved our understanding of the physical world. Many physics laws are described by systems of highly nonlinear differential equations that have direct implications for understanding and predicting physical dynamics. However, these equations are usually too complicated to be solvable. The current paradigm of numerical methods for solution approximation is purely physics-based: known physical laws encoded in systems of coupled differential equations are solved over space and time via numerical differentiation and integration schemes \citep{Houska2012, hughes2012finite, lisitsa2012finite, Butcher1996numerical, McDonough2007Turbulence, Sagaut2006Turbulence}. However, these methods are tremendously computationally intensive, requiring significant computational resources and expertise. An alternative is seeking simplified models that are based on certain  assumptions and roughly can describe the dynamics, such as Reynolds-averaged Navier-stokes equations for turbulent flows and Euler equations for gas dynamics \citep{Chaoua2017RANS, Labourasse2004RANS, tompson2017accelerating}. But it is highly nontrivial to obtain a simplified model that can describe a phenomenon with satisfactory accuracy. More importantly, for many complex real-world  phenomena, only partial knowledge of their dynamics is known. The equations may not fully represent the true system states.

Deep Learning (DL) provides efficient alternatives to learn high-dimensional spatiotemporal dynamics from massive datasets. It achieves so by directly predicting the input-output mapping and bypassing numerical integration. Recent works have shown that DL can generate realistic predictions and significantly accelerate the simulation of physical dynamics relative to numerical solvers, from turbulence modeling to weather prediction \citep{Wang2020TF, Kochkov2021MachineLA, Kim2019DeepFA, Karthik2021physics, Kima2019turbulence}. This opens up new opportunities at the intersection of DL and physical sciences, such as molecular dynamics\citep{Shi2021Learning, simm2021symmetryaware}, epidemiology\citep{Wu2021DeepGLEAMAH}, cardiology\citep{Linial2021GenerativeOM, Wang2020AorticPF} and material science \citep{Liu2019MultiFidelityPN, Cang2017ImprovingDP}. 

Despite the tremendous progress, DL is purely data-driven by nature, which has many limitations. DL models still adhere to the fundamental rules of statistical inference. The nonlinear and chaotic nature of real-world dynamics poses significant challenges to existing DL frameworks. Without explicit constraints, DL models are prone to make physically implausible forecasts, violating the governing laws of physical systems. Additionally, DL models often struggle with generalization: models trained on one dataset cannot adapt properly to unseen scenarios with different distributions,  known as distribution shift. For dynamics learning, the distribution shift occurs not only because the dynamics are non-stationary and nonlinear, but also due to the changes in system parameters, such as external forces and boundary conditions. In a word, the current limitation of DL models for learning complex dynamics is their lack of ability to understand the system solely from data and cope with the distributional shifts that naturally occur.

Neither DL alone nor purely physics-based approaches can be considered sufficient for learning complex dynamical systems in scientific domains. Therefore, there is a growing need for integrating traditional physics-based approaches with DL models so that we can make the best of both types of approaches. There is already a vast amount of work about physics-guided DL \citep{Willard2020Survey, Weinan2019Integrating, Brunton2019MachineLF, Kutz2017DeepLI, Karthik2021physics, prabhat_nature_2019, Brunton2021Fluid}, but the focus on deep learning for dynamical systems is still nascent. Physics-guided DL offers a set of tools to blend these physical concepts such as differential equations and symmetry with deep neural networks. On one hand, these DL models offer great computational benefits over traditional numerical solvers. On the other hand, the physical constraints impose appropriate inductive biases on the DL models, leading to accurate simulation, scientifically valid predictions, reduced sample complexity, and guaranteed improvement in generalization to unknown environments.  

This survey paper aims to provide a structured overview of existing methodologies of incorporating prior physical knowledge into DL models for learning dynamical systems. The paper is organized as below. 
\begin{itemize}[leftmargin=*,itemsep=1pt]
    \item Section \ref{obj} describes the significance of physics-guided DL.
    \item Section \ref{prob} formulates the four main learning problems of physics-guided DL, including solving differential equations, dynamics forecasting, learning dynamics residuals, and equation discovery.
    \item Section \ref{sec:loss}$\sim$\ref{sec:symmetry} categorizes existing physics-guided DL approaches into four groups based on the way how physics and DL are combined. Each lead with a detailed review of recent work as a case study and further categorized based on objectives or model architecture. 
    \begin{itemize}[leftmargin=10pt,itemsep=1pt]%
    \item Section \ref{sec:loss}: \texttt{Physics-guided loss function}: prior physics knowledge is imposed as additional soft constraints in the loss function.
    \item Section \ref{sec:architecture}: \texttt{Physics-guided architecture design}: prior physics knowledge is strictly incorporated into the design of neural network modules. 
    \item Section \ref{sec:hybrid}: \texttt{Hybrid physics-DL models}: complete physics-based approaches are directly combined with DL models. 
    \item Section \ref{sec:symmetry}: \texttt{Invariant and equivariant DL models}: DL models are designed to respect the symmetries of a given physical system.  
    \end{itemize}
    \item Section \ref{dis} summarizes the challenges in this field and discusses the emerging opportunities for future research. 
\end{itemize}  


\section{Significance of Physics-Guided Deep Learning}\label{obj}
This subsection provides an overview of the motivations and significance of physics-guided DL for learning dynamical systems. By incorporating physical principles, governing equations, mathematical modeling, and domain knowledge into DL models, the rapidly growing field of physics-guided DL can potentially (1)
accelerate data simulation (2) build physically scientifically valid models (3) improve the generalizability of DL models (4) discover governing equations. 

\subsection{Accelerate Data Simulation.}\label{data_simulation}
Simulation is an important method of analyzing, optimizing, and designing real-world processes, which are easily verified, communicated, and understood. It serves as the surrogate modeling and digital twin and  provides valuable  insights into complex physical systems. Traditional physics-based simulations often rely on running numerical methods: known physical laws encoded in systems of coupled differential equations are solved over space and time via numerical differentiation and integration schemes \citep{Houska2012, lisitsa2012finite, Butcher1996numerical, McDonough2007Turbulence, Sagaut2006Turbulence}. Although the governing equations of many physical systems are known, finding approximate solutions using numerical algorithms and computers is still prohibitively expensive. Because the discretization step size is usually confined to be very small due to stability constraints when the dynamics are complex. Moreover, the performance of numerical methods can highly depend on the initial guesses of unknown parameters \citep{iserles2009first}.  Recently, DL has demonstrated great success in the automation, acceleration, and streamlining of highly compute-intensive workflows for science \citep{prabhat_nature_2019, tompson2017accelerating, Kochkov2021MachineLA}. 

Deep dynamics models can directly approximate high-dimensional spatiotemporal dynamics by directly forecasting the future states and bypassing numerical integration \cite{Wang2020symmetry, Bezenac2018Deep, SanchezGonzalez2020LearningTS, Pfaff2021LearningMS, SanchezGonzalez2020LearningTS, Wang2021DyAd, pathak2022fourcastnet, li2021fourier}. These models are trained to make forward predictions given the historic frames as input with one or more steps of supervision and can roll out up to hundreds of steps during inference. DL models are usually faster than classic numerical solvers by orders of magnitude since DL is able to take much larger space or time steps than classical solvers \citep{Pfaff2021LearningMS}.

Another common approach is that deep neural networks can directly approximate the solution of complex coupled differential equations via gradient-based optimization, which is the so-called physics-informed neural networks (PINNs). This approach has shown success in approximating a variety of PDEs \citep{raissi2017physics, Raissi2018Hidden, Carleo2017SolvingTQ, Han2019SolvingMS}. Additionally, deep generative models, such as diffusion models and score-based generative models, have been shown effective in accurate molecule graph generation \citep{gnaneshwar2022score, hoogeboom2022equivariant}. The computer graphics community has also investigated using DL to speed up numerical simulations for generating realistic animations of fluids such as water and smoke \citep{Kim2019DeepFA, tompson2017accelerating, steffen2019latent, xie2018tempogan}. However, the community focuses more on the visual realism of the simulation rather than the physical characteristics.  


\subsection{Build Scientifically Valid Models.}
Despite the tremendous progress of DL for science, e.g., atmospheric science \citep{prabhat_nature_2019}, computational biology \citep{Alipanahi2015PredictingTS}, material science \citep{Cang2017ImprovingDP}, quantum chemistry \citep{Schtt2017SchNetAC}, it remains a grand challenge to incorporate physical principles in a systematic manner to the design, training, and inference of such models.  DL models are essentially statistical models that learn patterns from the data they are trained on. Without explicit constraints, DL models, when trained solely on data, are prone to make scientifically implausible predictions, violating the governing laws of physical systems. In many scientific applications, it is important that the predictions made by DL models are consistent with the known physical laws and constraints. For example, in fluid dynamics, a model that predicts the velocity field of a fluid must satisfy the conservation of mass and momentum. In materials science, a model that predicts the properties of a material must obey the laws of thermodynamics and the principles of quantum mechanics.

Thus, to build trustworthy predictive models for science and engineering, we need to leverage  known physical principles to guide DL models to learn the correct underlying dynamics instead of simply fitting the observed data. For instance, \cite{Anuj2017PGNN, jia2019physics, jinlong_2019, constrain1, Greydanus2019HamiltonianNN, Cranmer2020LagrangianNN} improve the physical and statistical consistency of DL models by explicitly regularising the loss function with physical constraints. Hybrid DL models, e.g., \citep{HybridNet, ayed2019learning, chen19} integrate differential equations in DL for temporal dynamics forecasting and achieve promising performance.  \cite{ling2016ra} and \cite{fang2018deep} studied tensor invariant neural networks that can learn the Reynolds stress tensor while preserving Galilean invariance. \cite{Wang2020TF} presented a hybrid model that combines the numerical RANS-LES coupling method with a custom-designed U-net. The model uses the temporal and spatial filters in the RANS-LES coupling method to guide the U-net in learning both large and small eddies. This approach improves the both accuracy and physical consistency of the model, making it more effective at representing the complex flow phenomena observed in many fluid dynamics applications.

\subsection{Improve the generalizability of DL models}
DL models often struggle with generalization: models trained on one dataset cannot adapt properly to unseen scenarios with distributional shifts that may naturally occur in dynamical systems \citep{Kouw2018domain, pan2010survey, long2015learning, Amodei2016Safety, Wang2020ODE}. Because they learn to represent the statistical patterns in the training data, rather than the underlying causal relationships. In addition, most current approaches are still trained to model a specific system and multiple systems with close distributions, making it challenging to meet the needs of the scientific domain with heterogeneous environments. Thus, it is imperative to develop generalizable DL models that can learn and generalize well across systems with various parameter domains.

Prior physical knowledge can be considered as an inductive bias that can place a prior distribution on the model class and shrink the model parameter search space. With the guide of inductive bias, DL models can better capture the underlying dynamics from the data that are consistent with physical laws. Across different data domains and systems, the laws of physics stay constant. Hence, integrating physical laws in DL enables the models to generalize  outside of the training domain and even to different systems. 

Embedding symmetries into DL models is one way to improve the generalization, which we will discuss in detail in subsection \ref{sec:symmetry}. For example, \cite{Wang2020symmetry} designed deep equivariant dynamics models that respect the rotation, scaling, and uniform motion symmetries in fluid dynamics. The models are both theoretically and experimentally robust to distributional shifts by symmetry group transformations and enjoy favorable sample complexity compared with data augmentation.
There are many other ways to improve the generalization of DL models by incorporating other physical knowledge. 
\cite{Wang2021DyAd} proposed a meta-learning framework to forecast systems with different parameters. It leverages prior
physics knowledge to distinguish different systems. Specifically, it uses an encoder to infer the physical parameters of
different systems and a prediction network to adapt and forecast giving the inferred system. Moreover, \cite{Erichson2019PhysicsinformedAF} encodes Lyapunov stability into an autoencoder model for predicting fluid flow and sea surface temperature. They show improved generalizability and reduced prediction uncertainty for neural nets that preserve Lyapunov stability. \cite{Shi2019Neural} shows adding spectral normalization to DNN to regularize its Lipschitz continuity can greatly improve the generalization to new input domains on the task of drone landing control.

\subsection{Discover Governing Equations}
One of the main objectives of science is to discover fundamental laws that can solve practical problems \cite{Weinan2019Integrating, dzeroski2007computational}. The discovery of governing equations is crucial as it enables us to comprehend the underlying physical laws that regulate complex systems. By identifying the mathematical models that describe the behavior of a system, we can make accurate predictions and gain insights into how the system will behave under different conditions. This knowledge can be applied to optimize the performance of engineering systems, improve the precision of weather forecasts, and understand the mechanisms behind biological processes, among other applications \cite{dvzeroski2007computational, brence2021probabilistic}. However, discovering governing equations is a challenging task for various reasons. Firstly, real-world systems are frequently complex and involve many interdependent variables, making it difficult to identify the relevant variables and their relationships. Secondly, many systems are nonlinear and involve interactions between variables that are hard to model using linear equations. Thirdly, the available data may be noisy or incomplete, making it challenging to extract meaningful patterns and relationships. Despite these challenges, recent advances in machine learning have made it possible to automate the process of governing equations discovery and identify complex, nonlinear models from data. These approaches may lead to new discoveries and insights into the behavior of complex systems for a wide range of applications.

Discovering governing equations from data is often accomplished by defining a large set of possible mathematical basis functions and learning the coefficients. \cite{Brunton2015Sparse, brunton2016discovering, kaiser2018sparse, brence2021probabilistic, schaeffer2017learning} proposed to find ordinary differential equations by creating a dictionary of possible basis functions and discovering sparse, low-dimensional, and nonlinear models from data using the sparse identification. More recent work, such as \cite{Lagergren2020LearningPD, Rudy2016pde}, incorporated neural networks to further augment the dictionary to model more complex dynamics.  \cite{Carderera2021CINDy} contributed to this trend by introducing an efficient first-order conditional gradient algorithm for solving the optimization problem of finding the best sparse fit to observational data in a large library of potential nonlinear models. Alternatively, \cite{martius2016extrapolation, sahoo2018learning} presented a shallow neural network approach, \textit{EQL} to identify concise equations from data. They replaced the activation functions with predefined basis functions, including identity and trigonometry functions, and used specially designed division units to model division relationships in the potential governing equations. Similarly, \cite{long2018pde, long2019pde} designed \textit{PDE-Nets} that use convolution to approximate differential operators and symbolic neural networks to approximate and recover multivariate functions. These models could learn various functional relations, with and without divisions, from noisy data in a confined domain. However, scalability, overfitting, and over-reliance on high-quality measurement data remain critical concerns in this research area \citep{rao2022discovering}.

\section{Problem Formulation}\label{prob}
In light of the motivation and significance of physics-guided deep learning we discuss in the previous section, the primary research efforts in this field have been aimed at tackling the following four fundamental problems.
\subsection{Solving Differential Equations} \label{prob:solving}
When $\mathcal{F}$ in Eq. \ref{eqn:dynamics} is \textit{known} but Eq. \ref{eqn:dynamics} is too complicated to be solvable, researchers tend to directly solve the differential Eq.ations by approximating solution of $\bm{u}(x)$ with a deep neural network, and enforcing the governing equations as a soft constraint on the output of the neural nets during training at the same time\citep{mazier1, raissi2017physics, krishnapriyan2021characterizing}. This approach can be formulated as the following optimization problem, 
\begin{equation}\label{equ:pinn_loss}
    \text{min}_{\theta} \; \mathcal{L}(\bm{u}) + \lambda_\mathcal{F} \mathcal{L}_\mathcal{F}(\bm{u})
\end{equation}
$\mathcal{L}(\bm{u})$ denotes the misfit of neural net predictions and the training data points. $\theta$ denotes the neural net parameters. $\mathcal{L}_\mathcal{F}(\bm{u})$ is a constraint on the residual of the differential equation system under consideration and $\lambda_\mathcal{F}$ is a regularization parameter that controls the emphasis on this residual. The goal is then to train the neural nets to minimize the loss function in Eq. \ref{equ:pinn_loss}.

\subsection{Learning Dynamics Residuals}
When $\mathcal{F}$ in Eq. \ref{eqn:dynamics} is \textit{partially know}, we can use neural nets to learn the errors or residuals made by physics-based models \citep{BelbutePeres2020CombiningDP, yin2021augmenting, Kani2017DRRNNAD}. The key is to learn the bias of physics-based models and correct it with the help of deep learning. The final prediction of the state is composed of the simulation from the physics-based models and the residual prediction from neural nets as below, 
\begin{equation}
\hat{\bm{u}} = \hat{\bm{u}}_\mathcal{F} + \hat{\bm{u}}_{\text{NN}}.
\end{equation}
where $\hat{u}_\mathcal{F}$ is the prediction obtained by numerically solving $\mathcal{F}$, $\hat{u}_{\text{NN}}$is the prediction from neural networks and $\hat{u}$ is the final prediction made by hybrid physics-DL models. 

This learning problem generally involves two training strategies: 1) joint training: optimizing the parameters in the differential equations and the neural networks at the same time by minimizing the prediction errors of the system states. 2) two-stage training: we first fit differential equations on the training data and obtain the residuals, then directly optimize the neural nets on predicting the residuals.

\subsection{Dynamics Forecasting}
When $\mathcal{F}$ in Eq. \ref{eqn:dynamics} is \textit{unknown} or numerically solving Eq. \ref{eqn:dynamics} requires too much computation, many works studied learning high-dimensional spatiotemporal dynamics by directly predicting the input-output system state mapping and bypassing numerical discretization and integration \citep{Bezenac2018Deep, Karthik2021physics,Shi2021Learning, Wang2020symmetry}. If we assume the first dimension $x_1$ of $\bm{u}$ in Eq. \ref{eqn:dynamics} is the time dimension $t$, then the problem of dynamics forecasting can be defined as learning a map $f: \mathbb{R}^{n \times k} \mapsto \mathbb{R}^{n \times q} $ that maps a sequence of historic states to future states of the dynamical system,
\begin{equation}\label{equ_task}
f(\bm{u} \text{\footnotesize $(t-k+1, \cdot)$}, ..., \bm{u}\text{\footnotesize $(t, \cdot)$}) = \bm{u}\text{\footnotesize $(t+1, \cdot)$}, ..., \bm{u}\text{\footnotesize $(t+q, \cdot)$}
\end{equation}
where $k$ is the input length and $q$ is the output length. $f$ is commonly approximated with purely data-driven or physics-guided neural nets and the neural nets are optimized by minimizing the prediction errors of the state $\mathcal{L}(\bm{u})$.

\subsection{Search for Governing Equations}
When $\mathcal{F}$ in Eq. \ref{eqn:dynamics} is \textit{unknown} and it is necessary to determine the precise governing equations to solve practical problems, numerous efforts have been made to discover the exact mathematical formulation of $\mathcal{F}$. The most common approach is to select from a wide range of possible candidate functions and choose the model that minimizes fitting errors on observation data.

More specifically, based on Definition \ref{dfn:dynamics}, the goal of discovering governing equations is to find an approximate function $\mathcal{\hat{F}} = \Phi(\bm{u}(x), x) \boldsymbol{\theta} \approx \mathcal{F}$, where $\Phi(\bm{u}(x), x) = [\phi_1(\bm{u}(x), x), \phi_2(\bm{u}, x), \ldots, \phi_p(\bm{u}, x)]$ is a library of candidate functions, such as polynomials and trigonometric functions, and $\boldsymbol{\theta} \in \mathbb{R}^p$ is a sparse vector indicating which candidate functions are active in the dynamics. This problem can be formulated as an optimization problem, where we aim to minimize the following cost function over a set of observed data $\{\bm{y}_i\}_{i=1}^n$ of $\bm{u}$:
\begin{equation}\label{equ_discovery}
\mathcal{L}(\boldsymbol{\theta}) = \sum_{i=1}^n ||\Phi(\bm{y}_i, x) \boldsymbol{\theta}||^2
\end{equation}

\section{Physics-Guided Loss Functions and Regularization}\label{sec:loss}
Complex physical dynamics occur over a wide range of spatial and temporal scales. Standard DL models may simply fit the observed data while failing to learn the correct underlying dynamics, thus leading to low physical consistency and poor generalizability.  One of the simplest and most widely used approaches to incorporate physical constraints is via designing  loss functions (regularization). Physics-guided loss functions (regularization) can assist DL models to capture correct and generalizable dynamic patterns that are consistent with physical laws. Furthermore, the loss functions constrained by physics laws can reduce the possible search space of parameters. This approach is sometimes referred to as imposing differentiable “soft” constraints, which will be contrasted with imposing “hard” constraints (physics-guided architecture) in the next section.  In this chapter, we will start with a case study of physics-guided loss functions, and then categorize these types of methods based on their objectives, including solving differential equations, improving prediction, and accelerating data generation. 


\subsection{Case Study: Physics-informed Neural Networks}
The Physics-informed Neural Networks (PINNs) approach \cite{Raissi2018Hidden, mazier1, cuomo2022scientific, raissi2017physics, cai2021physics} is a prime example of incorporating physics knowledge into the design of loss functions. PINNs have shown efficiency and accuracy in learning simple differential equations. Using fully connected neural networks, PINNs directly approximate the solution of differential equations with space coordinates and time stamps as inputs. These networks are trained by minimizing both the loss on measurements and the residual function error through the partial differential equation. More specifically, based on the Def. \ref{dfn:dynamics}, a fully connected neural network is employed to model solution $\bm{\hat{u}}(x, t | \bm{\theta}_{\text{PINN}})$, where $\bm{\theta}_{\text{PINN}}$ denotes the weights of the PINN  and be optimized by minimizing the following loss function. 
\begin{equation}
    \mathcal{L}_{\text{PINN}}=\mathcal{L}(\bm{u}) + \lambda_\mathcal{F} \mathcal{L}_\mathcal{F}(\bm{u})
\end{equation}

$\mathcal{L}(\bm{u})=\Vert \bm{\hat{u}}-\bm{y}\Vert _{\Gamma }$ is the error between the $\bm{\hat{u}}(x, t | \bm{\theta}_{\text{PINN}})$ and the set of boundary conditions and measured data on the set of points $\Gamma$ where the boundary conditions and data are defined. $\mathcal{L}_{\mathcal{F}}=\Vert \mathcal{F}(\bm{\hat{u}}(x, t|\bm{\theta}_{\text{PINN}}), x, t)\Vert _{\Gamma}$ is the mean-squared error of the residual function to enforce the predictions generated by PINNs satisfy the desired differential equations. 

However, while PINNs have shown some success in capturing the underlying physical phenomena, \cite{krishnapriyan2021characterizing} has pointed out that they often struggle to learn complex physical systems due to the difficulties posed by PDE regularizations in the optimization problem. Furthermore, the effectiveness of PINNs is highly dependent on the quality of the input data, and performance may suffer in the presence of noise or limited data \cite{chen2021physics, bajaj2021robust, yang2021b}.  Moreover, limited by poor generalizability of neural networks, PINNs have trouble generalizing to the space and time domain that is not covered in the training set \cite{Kouw2018domain, pan2010survey, Amodei2016Safety}. These limitations present significant challenges for the development and application of PINNs in real-world applications. Nonetheless, continued research into PINNs  may help to overcome these challenges and improve their ability to capture and predict complex physical phenomena.

\subsection{Solving Differential Equations} 
Continuing from the previous discussion of PINNs in the previous section, to overcome the optimization difficulties of PINNs, \cite{krishnapriyan2021characterizing} proposed two ways to alleviate this optimization problem. One is to start by training the PINN on a small constraint coefficient and then gradually increase the coefficient instead of using a big coefficient right away. The other one is training the PINN to predict the solution one time step at a time instead of the entire space-time at once. Furthermore, \cite{chen2021physics} found that PINNs can overfit and propagate errors on domain boundaries, even when using physics-inspired regularizers. To address this, they introduced Gaussian Process-based smoothing on boundary conditions to recover PINNs' performance against noise and errors in measurements. Moreover, \cite{yang2021b} proposed a Bayesian framework that combines PINNs with a Bayesian network. Compared to PINNs, the hybrid model can provide uncertainty quantification and more accurate predictions in scenarios with large noise because it can avoid overfitting. Apart from PINNs, \citep{Zhu2019PhysicsConstrainedDL} proposed to use flow-based generative models to learn the solutions of probabilistic PDEs while the PDE constraints are enforced in the loss function. \citep{Wang2020TF} investigated using neural nets to learn the evolution of the velocity fields of incompressible turbulent flow, the divergence of which is always zero. It found that constraining the model with a divergence-free regularizer can reduce the divergence of prediction and improve prediction accuracy.

\subsection{Improving Prediction Performance} 
Physics-guided loss functions or regularization have shown great success in improving prediction performance, especially the physical consistency of DL models.  
\cite{Anuj2017PGNN} used neural nets to model lake temperature at different times and different depths. They ensure that the predictions are physically meaningful by regularizing that the denser water predictions are at lower depths than predictions of less dense water. 
\cite{jia2019physics} further introduced a loss term that ensures thermal energy conservation between incoming and outgoing heat fluxes for modeling lake temperature.  
\cite{Beucler2019AchievingCO} designed conservation layers to strictly enforce conservation laws in their NN emulator of atmospheric convection.
\cite{Beucler2019EnforcingAC} introduced a more systematic way of enforcing nonlinear analytic constraints in neural networks via constraints in the loss function. 
\cite{Zhang2018DeepPM} incorporated the loss of atomic force and atomic energy into neural nets for improved accuracy of simulating molecular dynamics.
\cite{Liu2019MultiFidelityPN} proposed a novel multi-fidelity physics-constrained neural network for material modeling, in which the neural net was constrained by the losses caused by the violations of the model, initial conditions, and boundary conditions.  \cite{dona2021pdedriven} proposed a novel paradigm for spatiotemporal dynamics forecasting that performs spatiotemporal disentanglement using the functional variable separation. The specific-designed time invariance and regression loss functions ensure the separation of spatial and temporal information. 

Hamiltonian mechanics is a mathematical framework that describes the dynamics of a system in terms of the total energy of the system, which is the sum of the kinetic and potential energy. \cite{Greydanus2019HamiltonianNN} proposed \textit{Hamiltonian Neural Nets (HNN)} that parameterizes a Hamiltonian with a neural network and then learn it directly from data. The conservation of desired quantities is constrained in the loss function during training. The proposed \textit{HNN} has shown success in predicting mass-spring and pendulum systems. 
Lagrangian mechanics describes the dynamics of a system in terms of the difference between the kinetic energy and the potential energy of the system. \cite{Cranmer2020LagrangianNN} proposed \textit{Lagrangian Neural Nets (LNN)} used a neural network to parameterize the Lagrangian function that is the kinetic energy minus the potential energy. They trained the neural network with the Euler-Lagrange constraint loss functions such that it can learn to approximately conserve the total energy of the system. \cite{Finzi2020SimplifyingHA} further simplify the \textit{HNN} and \textit{LNN} via explicit constraints.
\cite{lee2021identifying} further introduced a meta-learning approach in \textit{HNN} to find the structure of the Hamiltonian that can be adapted quickly to a new instance of a physical system.
\cite{Yaofeng2020Benchmarking} benchmark recent energy-conserving neural network models based on Lagrangian/Hamiltonian dynamics on four different physical systems. 

\subsection{Data Generation} 
Simulation is an important method of analyzing, optimizing and designing real-world processes. Current numerical methods require significant computational resources when solving chaotic and complex differential equations. Because numerical discretization step size is confined to be very small due to stability constraints \citep{iserles2009first}. Also, the estimation of unknown parameters by fitting equations to the observed data requires much manual engineering in each application since the optimization of the unknown parameters in the system highly depends on the initial guesses. Thus, there is an increasing interest in utilizing deep generative models for simulating complex physical dynamics. Many works also imposed physical constraints in the loss function for better physical consistency. 

For instance, \cite{jinlong_2019} enforced the constraints of covariance into standard Generative Adversarial Networks (GAN) via statistical regularization, which leads to faster training and better physical consistency compared with standard GAN. 
\cite{xie2018tempogan} proposed \textit{tempoGAN} for super-resolution fluid flow, in which an advection difference loss is used to enforce the temporal coherence of fluid simulation. 
\cite{White2019DownscalingNW} modified \textit{ESRGAN}, which is a conditional GAN designed for super-resolution, by replacing the adversarial loss with a loss that penalizes errors in the energy spectrum between the generated images and the ground truth data. 
Conditional GAN is also applied to emulating numeric hydroclimate models in \cite{Manepalli2019EmulatingNH}. The simulation performance is further improved by penalizing the snow water equivalent via the loss function.
\cite{Kim2019DeepFA} proposed a generative model to simulate fluid flows, in which a novel stream function-based loss function is designed to ensure divergence-free motion for incompressible flows. \cite{Gao2020SuperresolutionAD} proposed a physics-informed convolutional model for flow super-resolution, in which the physical consistency of the generated high-resolution flow fields is improved by minimizing the residuals of Navier-Stokes equations.

\subsection{Pros and Cons}
While physics-guided loss functions are easy to design and use, and can improve prediction accuracy and physical consistency, they do have several limitations. Firstly, the physical constraints incorporated in the loss functions are usually considered soft constraints, and may not be strictly enforced. This means that the desired physical properties may not be guaranteed when the models are applied to new datasets. Secondly, PDE regularization can make loss landscapes more complex and cause optimization issues that are difficult to address, as noted in \citep{krishnapriyan2021characterizing}. Finally, there may be a trade-off between prediction errors and physics-guided regularizers. For example, \citep{Wang2020TF} investigated incompressible turbulent flow prediction using neural nets and found that while constraining the model with a divergence-free regularizer can reduce the divergence of predictions, too much regularization may smooth out small eddies in the turbulence, resulting in a larger prediction error.

\section{Physics-Guided Design of Architecture}\label{sec:architecture}
While incorporating physical constraints as regularizers in the loss function can improve performance, DL is still used as a black box model in most cases. The modularity of neural networks offers opportunities for the design of novel neurons, layers, or blocks that encode specific physical properties.  The advantage of physics-guided NN architectures is that they can impose ``hard'' constraints that are strictly enforced, compared to the ``soft'' constraints described in the previous section. The ``soft'' constraints are much easier to design than hard constraints, yet not required to be strictly satisfied. DL models with physics-guided architectures have theoretically guaranteed properties, and hence are more interpretable and generalizable. In this chapter, we will start with a case study of \texttt{Turbulent-Flow Net (TF-Net)} that unifies a popular Computational Fluid Dynamics (CFD) technique and a custom-designed U-net. We further categorize other related methods based on their architectural design. 

\begin{figure}[!htb]
\centering
\includegraphics[width= 0.85\linewidth]{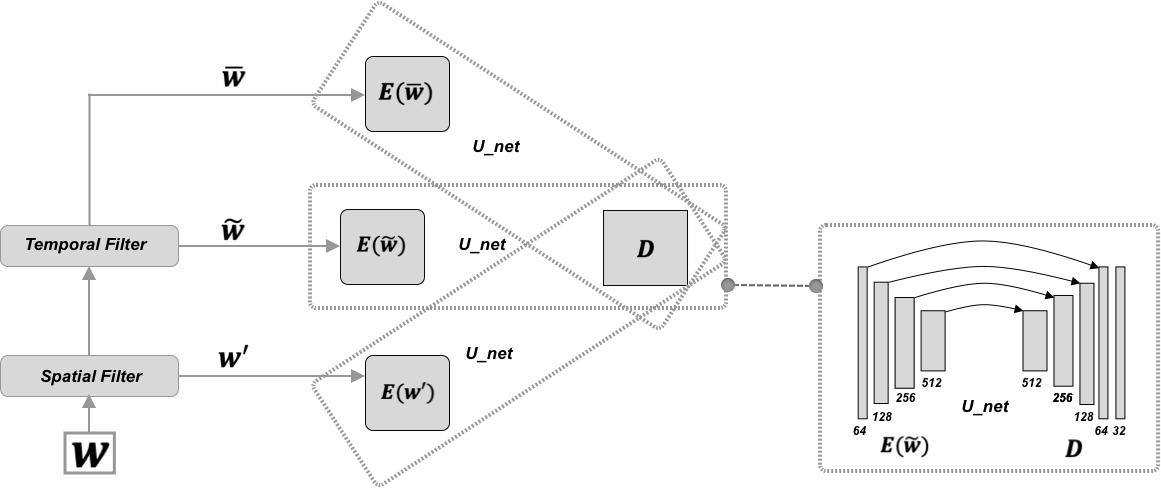}
    \caption{Turbulent Flow Net: three identical encoders to learn the transformations of the three components of different scales, and one shared decoder that learns the interactions among these three components to generate the predicted 2D velocity field at the next instant. Each encoder-decoder pair can be viewed as a U-net and the aggregation is a weighted summation.}
    \label{fig:tfnet}
\end{figure}

\subsection{Case Study: Turbulent-Flow Net}
\textit{TF-Net} \citep{Wang2020TF} is a physics-guided DL model for turbulent flow prediction. As shown in Figure \ref{fig:tfnet}, it applies scale separation to model different ranges of scales of the turbulent flow individually. Computational fluid dynamics (CFD) techniques are at the core of present-day turbulence simulation. Direct Numerical simulations (DNS) are accurate
but not computationally feasible for practical applications. Great emphasis was placed on the alternative approaches including Large Eddy Simulation (LES) and Reynolds-averaged Navier-Stokes
(RANS). Both resort to resolving large scales while modeling small scales, using various averaging techniques and/or low-pass filtering of the governing equations \citep{McDonough2007Turbulence, Sagaut2006Turbulence}.

One of the widely used CFD techniques, the RANS-LES coupling approach \citep{RANS-LES},  combines both Reynolds-averaged Equations (RANs) and Large Eddy Simulation (LES) approaches in order to take advantage of both methods. Inspired by RANS-LES coupling, \textit{TF-Net} replaces a priori spectral filters with trainable convolutional layers. The turbulent flow is decomposed into three components, each of which is approximated by a specialized U-net to preserve the multiscale properties of the flow. A shared decoder learns the interactions among these three components and generates the final prediction. The motivation for this design is to explicitly guide the ML model to learn the nonlinear dynamics of large-scale and Subgrid-Scale Modeling motions as relevant to the task of spatiotemporal prediction. In other words, we need to force the model to learn not only the large eddies but also the small ones. When we train a predictive model directly on the data with MSE loss, the model may overlook the small eddies and only focus on large eddies to achieve reasonably good accuracy.

\begin{figure*}[hbt!]
  \begin{minipage}[b]{0.3\textwidth}
     \includegraphics[width=\textwidth]{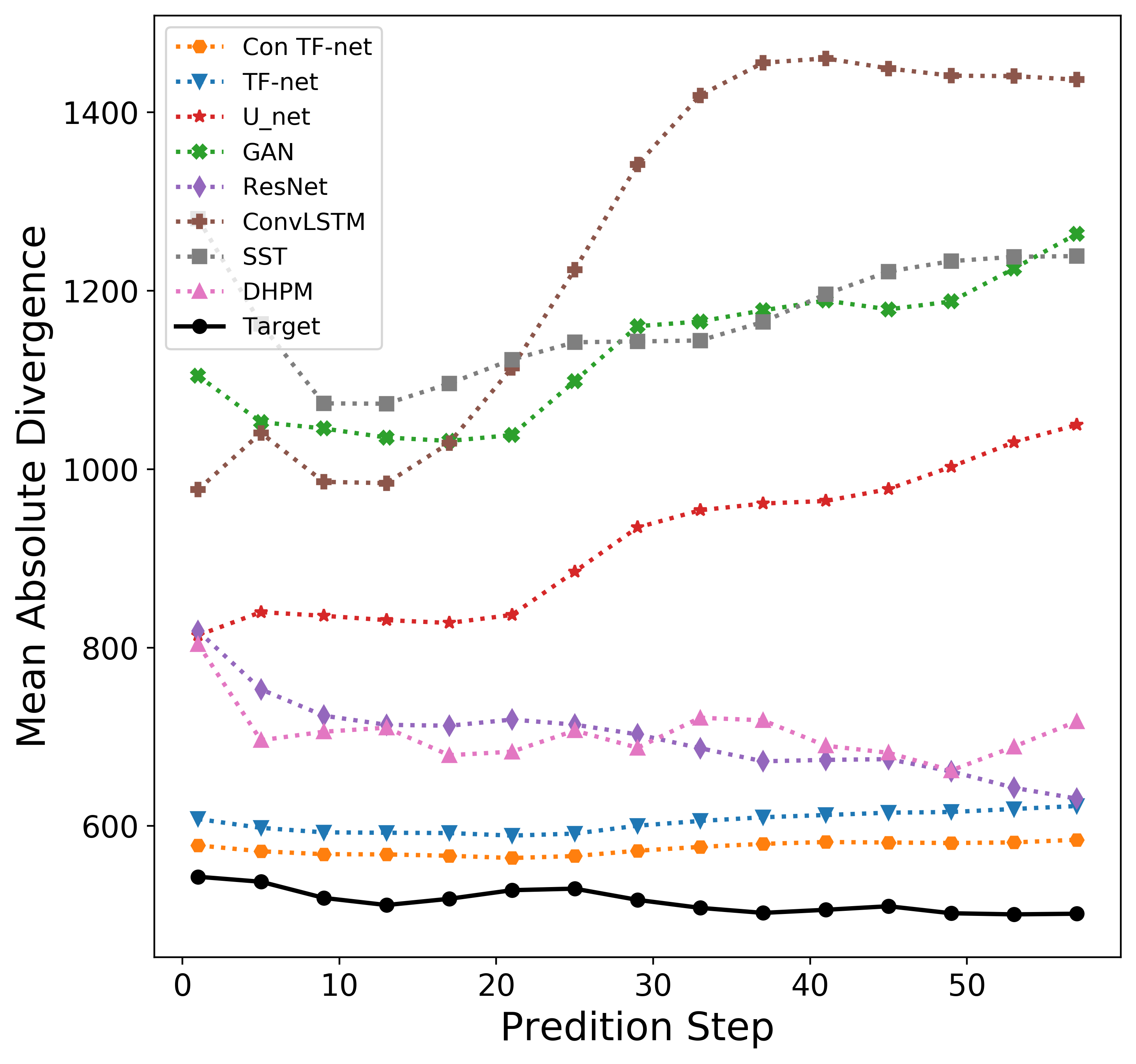}
  \end{minipage} 
  \begin{minipage}[b]{0.3\textwidth}
\includegraphics[width=\linewidth]{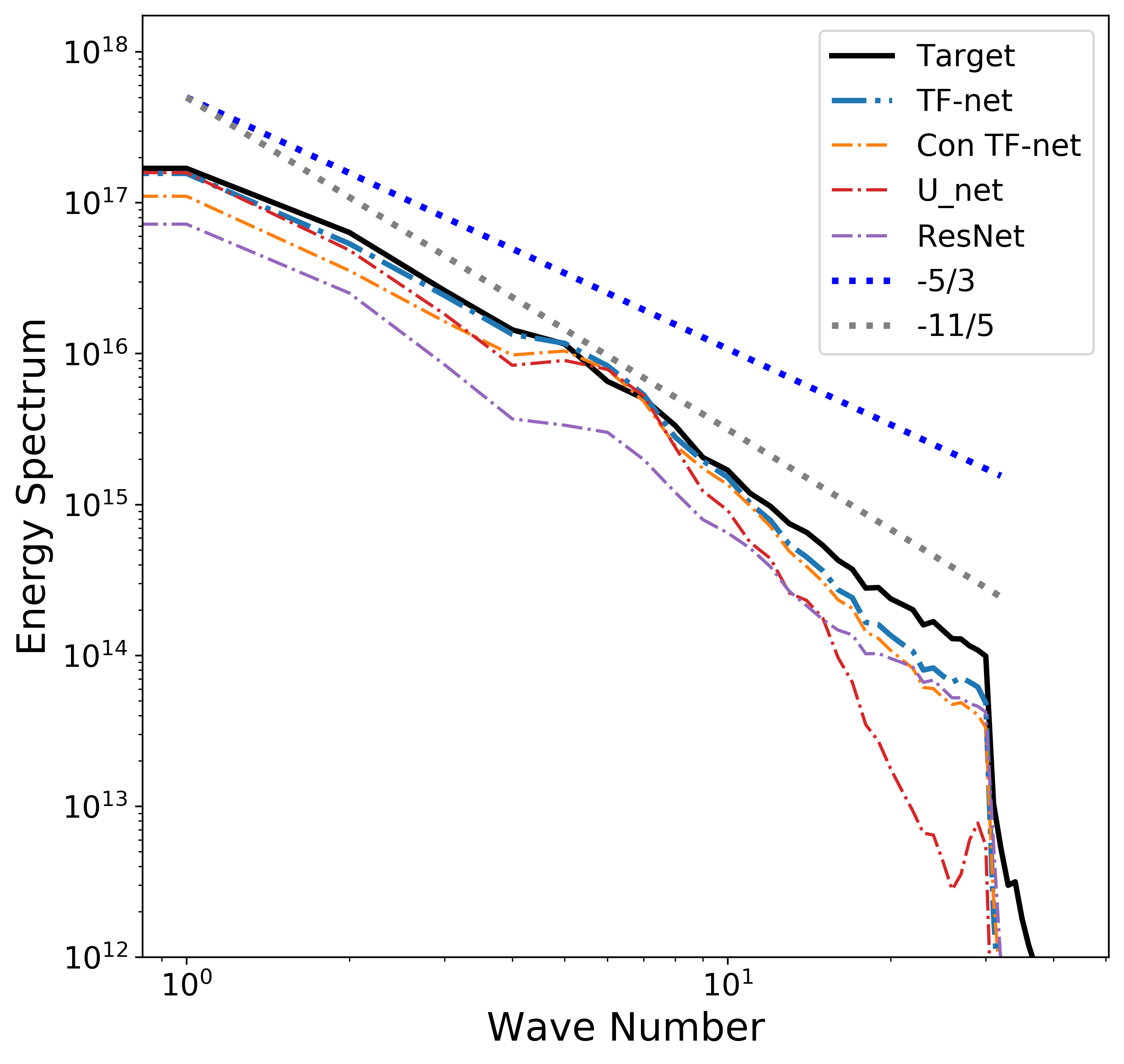}
  \end{minipage} 
    \begin{minipage}[b]{0.37\textwidth}
  \includegraphics[width= \linewidth]{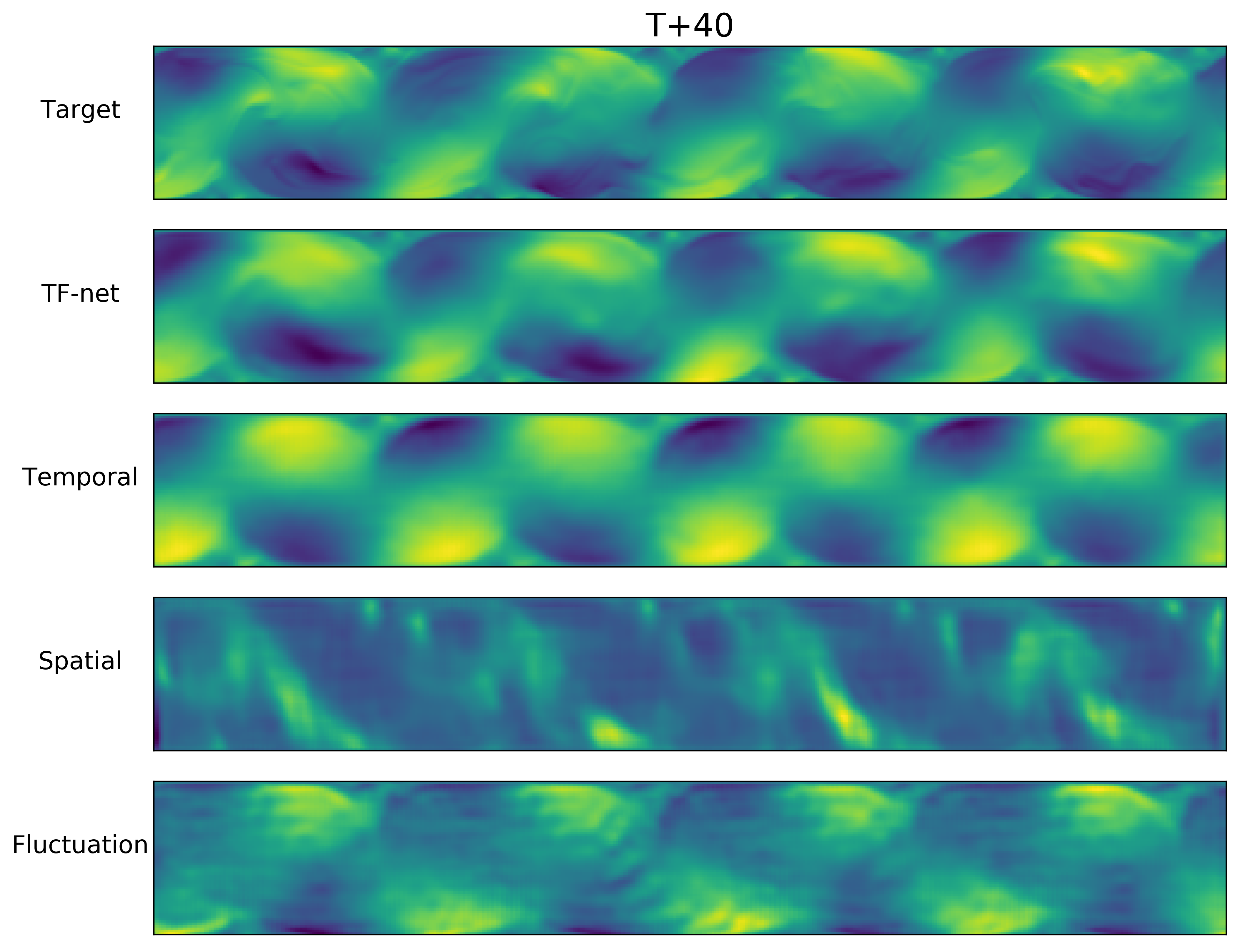}
 \end{minipage} 
 \caption{From left to right: Mean absolute divergence of different models' predictions at varying forecasting horizon; The Energy Spectrum of the target, \textit{TF-Net}, U-net and ResNet on the leftmost square sub-region; Ablation Study: Ground truth, prediction from \textit{TF-Net} and the outputs of each small U-net while the other two encoders are zeroed out.}
 \label{results}
\end{figure*}

\begin{figure*}[htb!]
\centering
  \begin{minipage}[b]{0.326\textwidth}
\includegraphics[width= \linewidth]{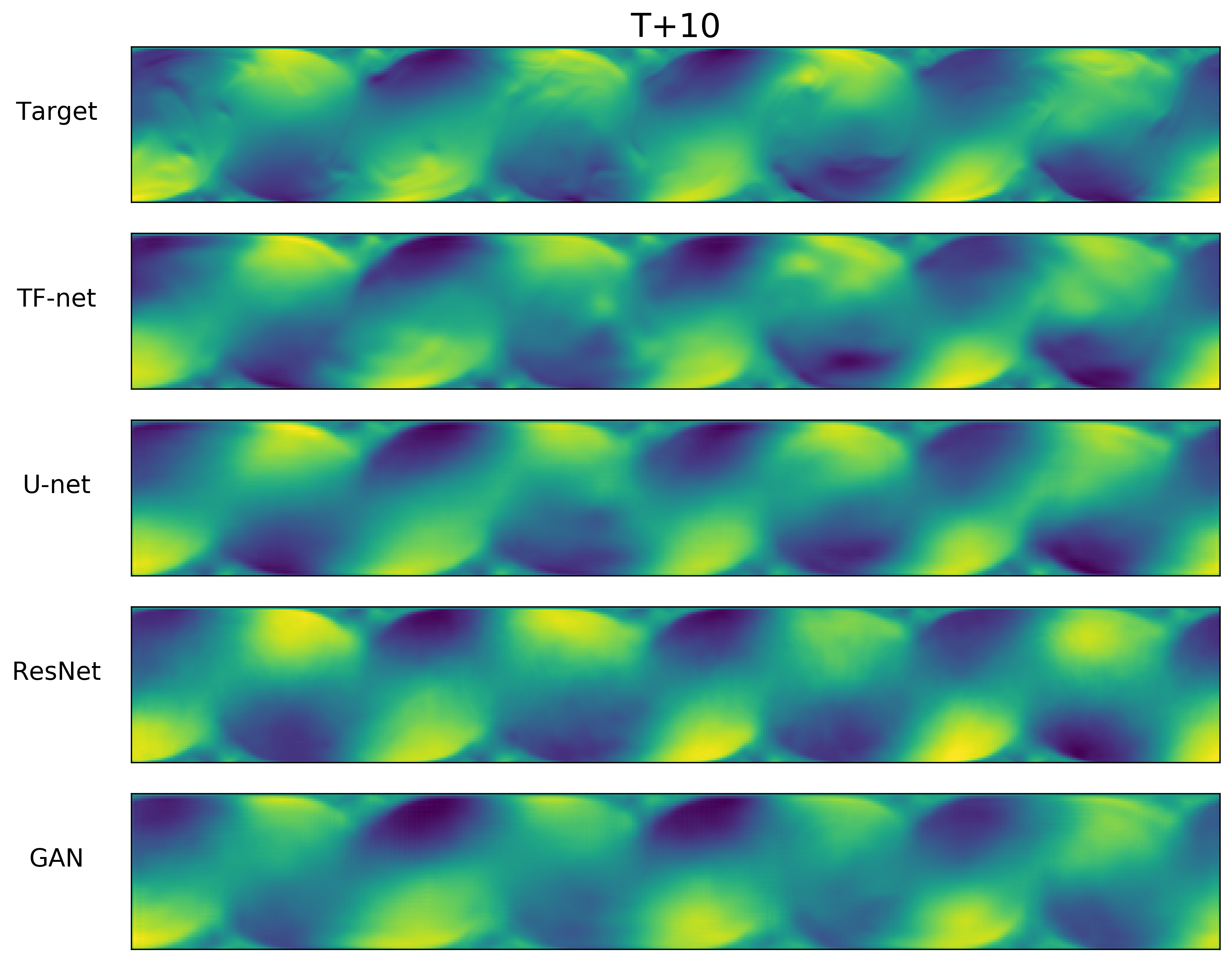}
  \end{minipage} 
\begin{minipage}[b]{0.326\textwidth}
\includegraphics[width= \linewidth]{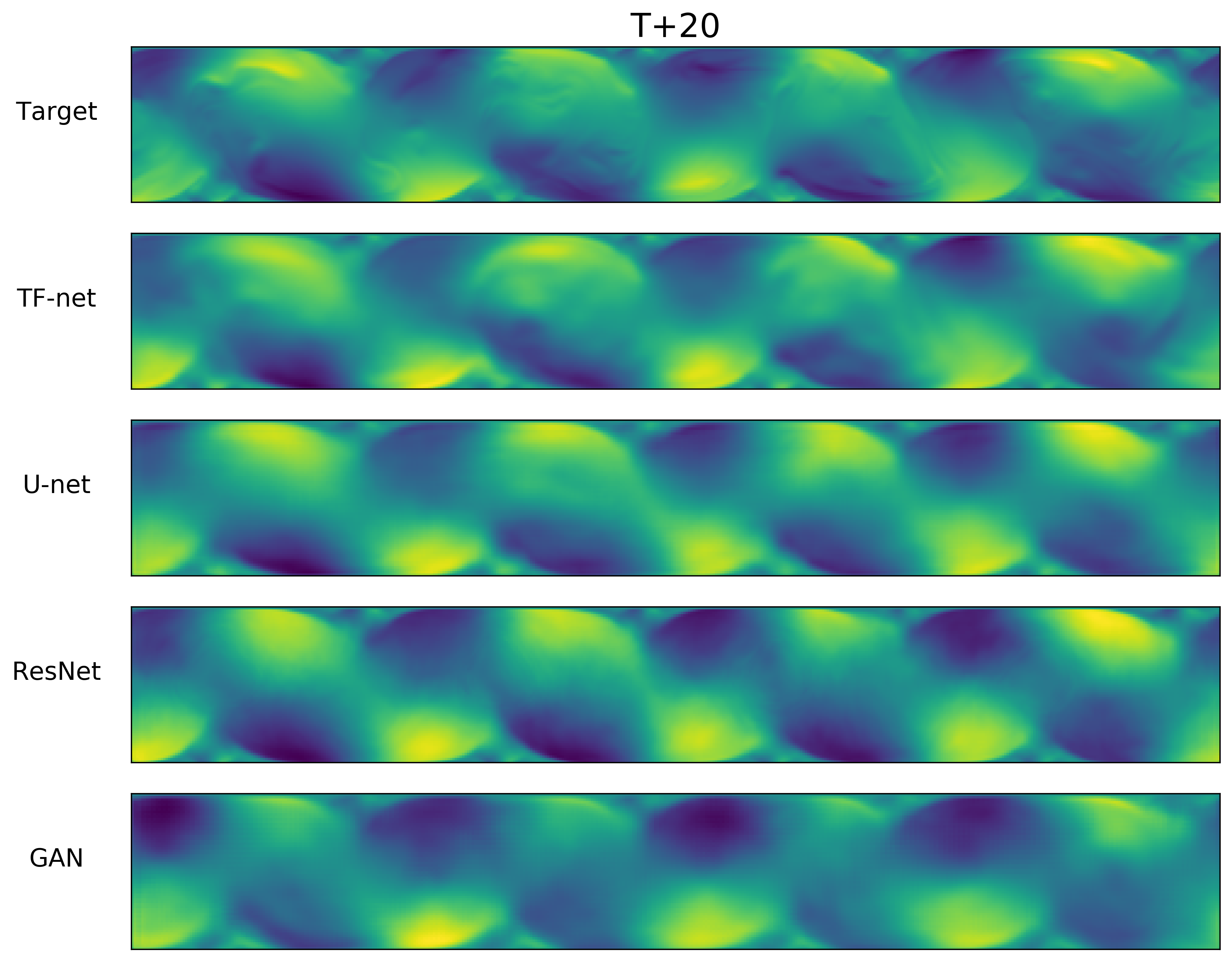}
  \end{minipage} 
\begin{minipage}[b]{0.326\textwidth}
\includegraphics[width= \linewidth]{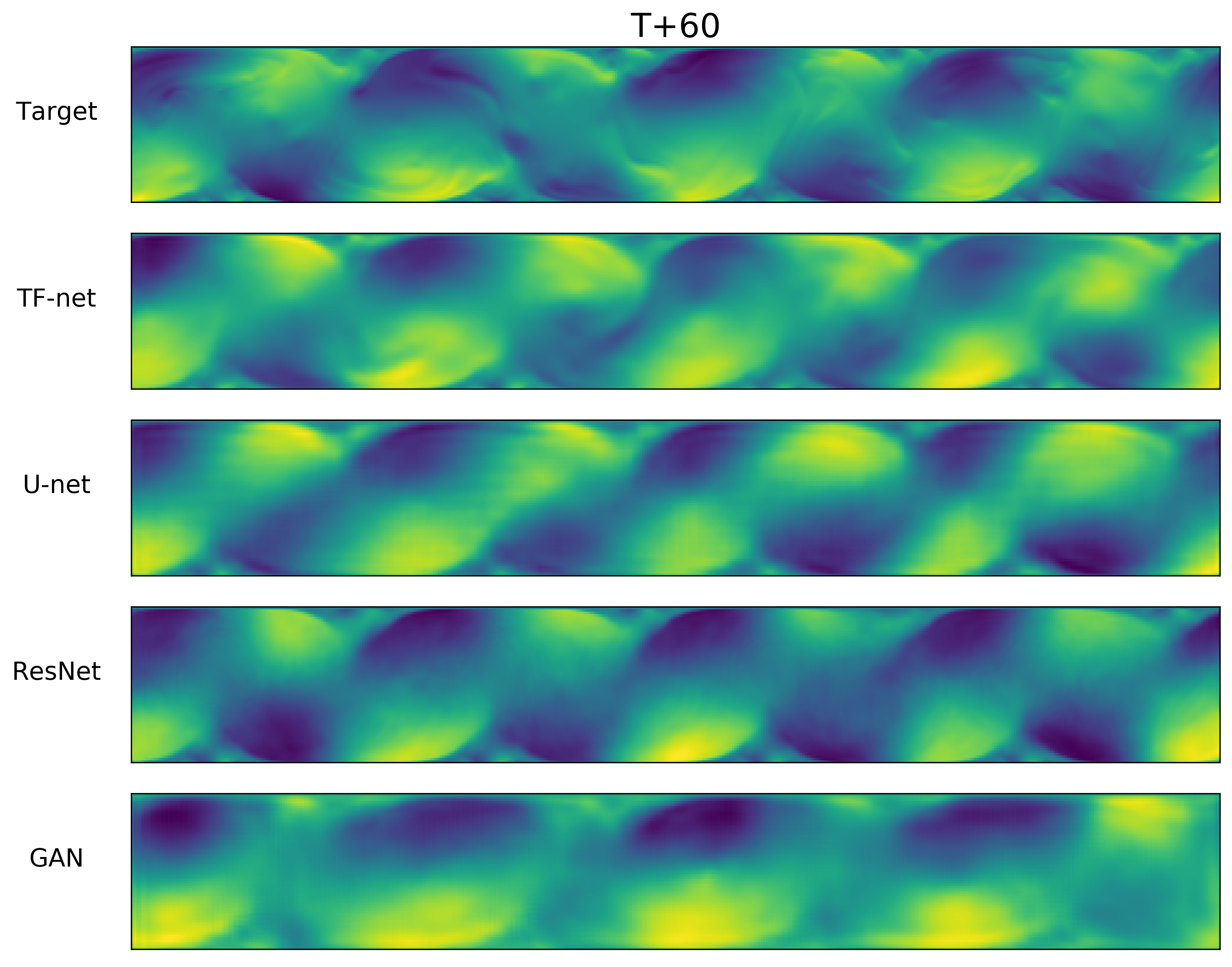}
  \end{minipage} 
\caption{Ground truth and predicted velocity $u$ by \textit{TF-Net} and three best baselines (U-Net, ResNet, and GAN) at time $T+10$, $T+30$ to $T+60$ (suppose $T$ is the time step of the last input frame).}
\label{rbc_pred}
\end{figure*}

Besides RMSE, physically relevant metrics including divergence and energy spectrum are used to evaluate the performance of the model's prediction. Figure \ref{results} shows \textit{TF-Net} consistently outperforms all baselines on physically relevant metrics (Divergence and Energy Spectrum). Constraining it with the divergence-free regularizer that we described in the previous section can further reduce the Divergence. Figure \ref{rbc_pred} shows the ground truth and predicted velocity along $x$ direction by \textit{TF-Net} and three best baselines. We see that the predictions by our TF-Net model are the closest to the target based on the shape and frequency of the motions. 

We also perform an ablation study to understand each component of TF-Net and investigate whether the model has actually learned the flow with different scales. Figure \ref{results} right includes  predictions and the outputs of each small U-net while the other two encoders are zeroed out. We observe that the outputs of each small U-net are the flow with different scales, which demonstrates that can learn multi-scale behaviors. In summary, \textit{TF-Net} is able to generate both accurate and physically meaningful predictions of the velocity fields that preserve critical quantities of relevance.

\subsection{Convolutional architecture}
Convolutional architecture remains dominant in most tasks of  computer vision, such as objection, image classification, and video prediction. Thanks to their efficiency and desired inductive biases, such as locality and translation equivariance, convolution neural nets have  been widely applied to emulate and predict complex spatiotemporal physical dynamics. Researchers have proposed various ways to bake desired physical properties into the design of convolutional models. 

For example, \cite{jiang2020enforcing} proposed to enforce hard linear spatial PDE constraints within CNNs using the Fast Fourier Transform algorithm.
\cite{Daw2020PhysicsGuidedA} modified the LSTM units to introduce an intermediate variable to  preserve monotonicity in a convolutional auto-encoder model for lake temperature.  
\cite{Muralidhar2020PhyNetPG} proposed a physics-guided convolutional model, \textit{PhyDNN}, which uses physics-guided structural priors and physics-guided aggregate supervision for modeling the drag forces acting on each particle in a computational fluid dynamics-discrete element Method. \cite{HybridNet} designed \textit{HybridNet} for dynamics predictions that combine ConvLSTM for predicting external forces with model-driven computation with CeNN for system dynamics. \textit{HybridNet} achieves higher accuracy on the tasks of forecasting heat convection-diffusion and fluid dynamics. 
\cite{Holl2020Learning} proposed to combine deep learning and a differentiable PDE solver for understanding and controlling complex nonlinear physical systems over a long time horizon. 
\cite{Schtt2017SchNetAC} proposed continuous-filter convolutional layers for modeling quantum interactions. The convolutional kernel is parametrized by neural nets that take relative positions between any two points as input. They obtained a joint model for the total energy and interatomic forces that follow fundamental quantum-chemical principles.

In addition, convolution layers have the potential to uncover governing equations. For instance, \cite{long2018pde, long2019pde}  developed \textit{PDE-Net}, which utilizes convolution to approximate differential operators over spatial domains of different orders. It also includes a symbolic neural network based on \textit{EQL} \cite{martius2016extrapolation, sahoo2018learning} to approximate and recover multivariate functions. The authors demonstrated that \textit{PDE-Net} is more compact than \textit{SINDy} dictionaries \cite{brunton2016discovering} and numerical experiments suggest that it can uncover the hidden PDE of various observed dynamics.

\subsection{Graph Neural Networks}
Standard convolutional neural nets only operate on regular or uniform mesh such as images. Graph neural networks move beyond data on the regular grid towards modeling objects with arbitrary positions. For instance, graph neural networks can model the fluid dynamics on irregular meshes that CNNs cannot. \cite{SanchezGonzalez2020LearningTS} designed a deep encoder-processor-decoder graphic architecture for simulating fluid dynamics under Lagrangian description. The rich physical states are represented by graphs of interacting particles, and complex interactions are approximated by learned message-passing among nodes. 

\cite{Pfaff2021LearningMS} utilized the same architecture to learn mesh-based simulation. The authors directly construct graphs on the irregular meshes constructed in the numerical simulation methods. In addition, they proposed an adaptive re-meshing algorithm that allows the model to accurately predict dynamics at both large and small scales. 
\cite{brandstetter2022message} further proposed two tricks to address the instability and error accumulation issues of training graph neural nets for solving PDEs. One is perturbing the input by a certain noise and only backpropagating errors on the last unroll step, and the other one is predicting multiple steps simultaneously in time. Both tricks make the model faster and more stable. 

\cite{li2020neural} proposed a \textit{Neural Operator} approach that learns the mapping between function spaces, and is invariant to different approximations and grids. More specifically, it used the message-passing graph network to learn Green's function from the data, and then the learned Green's function can be used to compute the final solution of PDEs. 
\cite{li2021fourier} further extended it to \textit{Fourier Neural Operator} by replacing the kernel integral operator with a convolution operator defined in Fourier space, which is much more efficient than \textit{Neural Operator}. In \cite{SanchezGonzalez2018GraphNA}, graph networks were also used to represent, learn, and infer robotics systems, bodies, and joints. \citep{Li2020Learning} proposed to learn compositional Koopman operators, using graph neural networks to encode the state into object-centric embeddings and using a block-wise linear transition matrix to regularize the shared structure across objects. Another important line of work is incorporating symmetries to design equivariant graph neural nets for modeling molecular dynamics, which will discuss in detail in Section \ref{sec:symmetry}.

\subsection{Multilayer Perceptron}
One of the main applications of Multilayer perceptron(MLP) in physics-guided architecture design is finding the linear Koopman operator. Koopman theory \citep{Koopman1931HamiltonianSA} provides a way to represent a nonlinear dynamical system using an infinite-dimensional linear Koopman operator that acts on a Hilbert space of measurement functions of the system state. However, finding the appropriate measurement functions that map the dynamics to the function space, as well as an approximate and finite-dimensional Koopman operator, is highly nontrivial. One way to obtain an approximation of the Koopman operator is through the Dynamic Mode Decomposition algorithm \citep{Schmid2008DynamicMD}, but this requires manually preparing nonlinear observables, which is not always feasible as prior knowledge about them may be lacking.

To address this challenge, recent research has explored using neural networks to learn the Koopman operator. One popular approach hypothesizes that there exists a data transformation that can be learned by neural networks, which yields an approximate finite-dimensional Koopman operator. For example, \cite{Yeung2019LearningDN} and \cite{Takeishi2017LearningKI} have proposed using fully connected neural networks to directly map the observed dynamics to a dictionary of nonlinear observables that span a Koopman invariant subspace. This mapping is represented through an autoencoder network, which embeds the observed dynamics onto a low-dimensional latent space where the Koopman operator is approximated by a linear layer. \cite{Lusch2018DeepLF} have further generalized this approach to enable learning the Koopman operator for systems with continuous spectra. \cite{Azencot2020ForecastingSD} have also designed a similar autoencoder architecture for forecasting physical processes. But in the latent space, the consistency of both the forward and backward systems is ensured, while other models only consider the forward system. This approach performs well on systems that have both forward and backward dynamics, enabling long time prediction.

Koopman theory can also be used to model real-world dynamics without known governing laws. \citep{Wang2023Koopman} have developed a novel approach, \textit{Koopman Neural Forecaster (KNF)}, to forecast highly non-stationary time series in an interpretable and robust manner. This approach uses Koopman theory to simplify non-linear real-world dynamics into linear systems, which then can be easily manipulated by modifying the Koopman matrix. It employs predefined measurement functions to impose appropriate inductive biases and uses a Koopman operator that varies over time to capture the underlying changing distribution. The model outperforms the state-of-the-art on highly non-stationary time series datasets, including M4, cryptocurrency return forecasting, and sports player trajectory prediction.

\subsection{Pros and Cons}
Embedding physics into the design of the model architecture can enable physical principles strictly enforced and theoretically guaranteed. That leads to more interpretable and generalizable deep learning models. However, it is not trivial to design physics-guided architectures that perform and generalize well without hurting the representation power of neural nets. Hard inductive biases can greatly improve the sample efficiency of learning, but could potentially become restrictive when the size of the dataset is big enough for models to learn all the necessary inductive biases from the data or when the inductive biases are not strict.

\section{Hybrid Physics-DL Model}\label{sec:hybrid}
The papers discussed in the previous two sections focus on incorporating the known properties of physical systems into the design of loss functions or neural network modules. In this section, we  talk about works that directly combine pure physics-based models, such as numerical methods, with DL models. 

\subsection{Case Study: Neural Differential Equations}
Neural Ordinary Differential Equations (\textit{Neural ODEs}) \cite{chen2018neural} generalize traditional RNNs that process data sequentially in discrete time steps by modeling data as continuous functions that change over time, allowing for a more flexible way to capture complex dynamics. They changed the traditionally discretized neuron layer depths into continuous equivalents such that the derivative of the hidden state can be parameterized using a neural network. The output of the network is then computed using a black box differential equation solver, making \textit{Neural ODEs} an efficient combination of neural nets and numerical solvers.

More specifically, they parametrize the velocity $\bm{\hat{z}}$ of a hidden state $\bm{z}$ with the
help of a neural network  $\bm{\hat{z}} = f_{\theta}(\bm{z}, t)$. Given the initial time $t_0$ and target time $t_T$, \textit{Neural ODEs} predict the target state $\bm{\hat{y}}_T$ by performing the following encoding, integration, and decoding operations:
 
\begin{equation}
    \bm{z}(t_0) = \phi_{\text{enc}}(\bm{y}_0), \;\;\;\;\;\; \bm{z}(t_T) = \bm{z}(t_0) + \int_{t_0}^{t_T} f_{\bm{\theta}}(\bm{z}, t) dt, \;\;\;\;\;\; \bm{\hat{y}}_T= \psi_{\text{dec}}(\bm{z}(t_T))
\end{equation}

where the encoder $\phi_{\text{enc}}$ and the decoder $\psi_{\text{dec}}$ can be neural networks. Solving an ODE numerically is commonly done by discretization and integration, such as the simple Euler method and higher-order variants of the Runge-Kutta method. However, all of these are computationally intensive since they require backpropagating through the operations of the solvers and store any intermediate quantities of the forward pass, incurring a high memory cost. Thus, the adjoint method \cite{pontryagin1987mathematical} is used to efficiently compute gradients during backpropagation. To compute the gradients of a loss function $L$ with respect to the initial state $\bm{z}(t_0)$ and the parameters $\bm{\theta}$, the key idea of the adjoint method is to introduce an adjoint state $\bm{p}(t)$, $\bm{p}(t) = \frac{\partial L}{\partial \bm{z}(t)}$, which satisfies the following differential equation:
\begin{equation}
\frac{d\bm{p}(t)}{dt} = -\bm{p}(t)^T \frac{\partial f_{\bm{\theta}}(\bm{z}(t), t)}{\partial \bm{z}} \label{adjoint_1}
\end{equation}
The adjoint state is used to compute the gradients of the loss function with respect to the initial state and the parameters using the following formulas:
\begin{equation}
 \frac{\partial L}{\partial \bm{\theta}} =  - \int_{t_T}^{t_0}\bm{p}(t)^T  \frac{\partial f_{\bm{\theta}}(\bm{z}(t), t)}{\partial \bm{\theta}} dt; \label{adjoint_2}
\end{equation}
In a word, these formulas can be computed efficiently by solving the ODE for $\bm{p}(t)$ using the same numerical method used to solve the forward ODE. During the forward pass, the ODE solver computes the solution of the differential equation $\bm{z}(t)$ using the initial state $\bm{z}(t_0)$ and the function $f_{\bm{\theta}}(\bm{z}(t), t)$. During the backward pass, the adjoint state $\bm{p}(t)$ is computed by solving Eqn. \ref{adjoint_1} that starts from the final time $t_T$ and backpropagates through time. This adjoint state is then used to compute the gradients of the loss function with respect to the initial state and the parameters of the ODE function in Eqn. \ref{adjoint_2}, which can then be used to update the model parameters through gradient descent.

\textit{Neural ODEs} have broad potential applications, particularly in domains that require continuous and dynamic models. They offer a useful tool for building continuous-time time series models, which can easily handle data coming at irregular intervals. They also allow for building normalizing flow, which makes it easy to track the change in density, even for unrestricted neural architecture. There have been several follow-up works that further extended the idea of continuous neural nets. For instance,  \cite{Dupont2019AugmentedNO} introduced \textit{Augmented Neural ODE} that is more expressive, empirically more stable, and lower computationally efficient than Neural ODEs. More importantly, it can learn the functions that have continuous trajectories mappings intersecting each other, which \textit{Neural ODEs} cannot represent.
\cite{Poli2019GraphNO} further extended this idea of continuous neural nets to graph convolutions, and proposed \textit{Graph Neural ODE}. \cite{Liu2019NeuralSS} proposed Neural Stochastic Differential Equation (\textit{Neural SDE}), which models stochastic noise injection by stochastic differential equations. They demonstrated that incorporating the noise injection regularization mechanism into the continuous neural network can reduce overfitting and achieve lower generalization errors. \cite{Linial2021GenerativeOM} proposed a Neural ODE-based generative time-series model that uses the known differential equation instead of treating it as hidden unit dynamics so that they can integrate mechanistic knowledge into the Neural ODE. \cite{Rackauckas2020UniversalDE} utilized neural networks to directly approximate the unknown terms in the differential equations. By using the adjoint method, the proposed model can efficiently compute gradients with respect to all parameters in the model, including the initial conditions, the parameters of the ODE, and the boundary conditions.

\subsection{Residual Modeling} 
Perhaps the simplest form of hybrid modeling  is residual learning, where DL learns to predict the errors or residuals made by physics-based models. The key is to learn the bias of physics-based models and correct it with the help of DL models \citep{Forssell1997COMBININGSA, Thompson1994ModelingCP}. A representative example is \textit{DeepGLEAM} \citep{Wu2021DeepGLEAMAH} for forecasting COVID-19 mortality that combines a mechanistic epidemic simulation model GLEAM with DL. It uses a Diffusion Convolutional RNN \citep{li2018dcrnn_traffic} (DCRNN) to learn the correction terms from GLEAM, which leads to improved performance over either purely mechanistic models or purely DL models on the task of one-week ahead COVID-19 death count predictions

Similarly, \cite{BelbutePeres2020CombiningDP} combines graph neural nets with a CFD simulator run on a coarse mesh to generate high-resolution fluid flow prediction. CNNs are used to correct the velocity field from the numerical solver on a coarse grid in \cite{Kochkov2021MachineLA}. \cite{Maulik2019SubgridMF} utilized neural networks for subgrid modeling of the LES simulation of two-dimensional turbulence. In \cite{San2018NeuralNC}, a neural network model is implemented in the reduced order modeling framework to compensate for the errors from the model reduction. \cite{Kani2017DRRNNAD} proposed \textit{DR-RNN} that is trained to find the residual minimizer of numerically discretized ODEs or PDEs. They showed that \textit{DR-RNN} can greatly reduce both computational cost and time discretization error of the reduced order modeling framework.  \cite{yin2021augmenting} introduced the \textit{APHYNITY} framework that can efficiently augment approximate physical models with deep data-driven networks. A key feature of their method is being able to decompose the problem in such a way that the data-driven model only models what cannot be captured by the physical model. 

\subsection{Intermediate Variable Modeling}
DL models can be used to replace one or more components of physics-based models that are difficult to compute or unknown. For example,  \cite{tompson2017accelerating} replaced the numerical solver for solving Poisson's equations with convolution networks in the procedure of Eulerian fluid simulation, and the obtained results are realistic and showed good generalization properties. \cite{Parish2016APF} proposed to use neural nets to reconstruct the model corrections in terms of variables that appear in the closure model. \cite{de2018deep} applied a U-net to estimate the velocity field given the historical temperature frames, then used the estimated velocity to forecast the sea surface temperature based on the closed-form solution of the advection-diffusion equation. \cite{Manzhos2006ARH} combined the high-dimensional model representation that is represented as a sum of mode terms each of which is a sum of component functions with NNs to build multidimensional potential, in which NNs are used to represent the component functions that minimize the error mode term by mode term.

\subsection{Pros and Cons}
Combining physics-based and deep learning models can enable leveraging both the flexibility of neural nets for modeling unknown parts of the dynamics and the interpretability and generalizability of physics-based models. However, one potential downside of hybrid physics-DL models worth mentioning is that all or most of the dynamics could be captured by neural nets and the physics-based models contribute little to the learning. 
That would hurt the interpretability and the generalizability of the model. We must ensure an optimal balance between the physics-based and DL models. We need neural nets to only model the information that cannot be represented by the physical prior.

\section{Invariant and Equivariant DL Models}\label{sec:symmetry}
Symmetry has long been implicitly used in DL to design networks with known invariances and equivariances.  Convolutional neural networks enabled breakthroughs in computer vision by leveraging translation equivariance \citep{zhang1988shift, lecun1989backpropagation, zhang1990parallel}. Similarly, recurrent neural networks \citep{rumelhart1986learning, hochreiter1997long}, graph neural networks \cite{scarselli2008graph, kipf2016semi}, and capsule networks \citep{sabour2017dynamic, hinton2011transforming} all impose symmetries.
While the equivariant DL models have achieved remarkable success in image and text data \citep{cohen2019gauge,  weiler2019e2cnn, cohen2016group, chidester2018rotation, Lenc2015understanding, kondor2018generalization, bao2019equivariant, worrall2017harmonic, cohen2016steerable, finzi2020generalizing, weiler2018learning, dieleman2016cyclic, Ghosh19Scale, Sosnovik2020Scale-Equivariant}, the study of equivariant nets in learning dynamical systems has become increasingly popular recently \cite{ling2016reynolds, Wang2020symmetry, smidt2021euclidean, hoogeboom2022equivariant, villar2021scalars, simm2021symmetryaware}. Since the symmetries can be integrated into neural nets through not only loss functions but also the design of neural net layers and there has been a large volume of works about equivariant and invariant DL models for physical dynamics, we discuss this topic separately in this section. 

\begin{wrapfigure}{r}{0.4\textwidth}
\centering
\includegraphics[width= 0.4\textwidth,trim={55 130 55 0}]{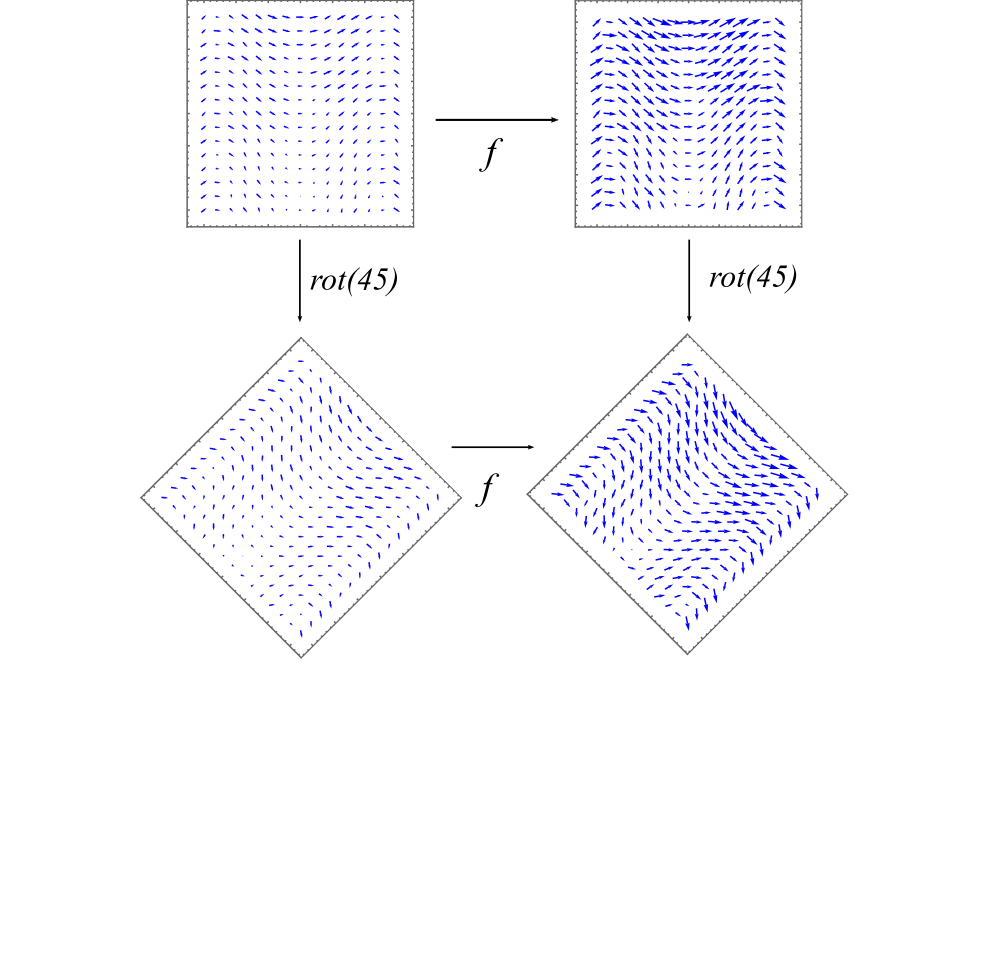} 
\caption{Illustration of equivariance: $f(x)=2x$ w.r.t $T = \mathrm{rot}(\pi/4)$}
\label{fig:equi}
\end{wrapfigure}

In physics, there is a deep connection between symmetries and physics. Noether’s law gives a correspondence between conserved quantities and groups of symmetries. For instance, translation symmetry corresponds to the conservation of energy and rotation symmetry corresponds to the conservation of angular momentum. By building a neural network that inherently respects a given symmetry, we thus make conservation of the associated quantity more likely and consequently the model’s prediction more physically accurate. Furthermore, by designing a model that is inherently equivariant to transformations of its inputs, we can guarantee that our model generalizes automatically across these transformations, making it robust to distributional shifts. 

A \textbf{group of symmetries} or simply \textbf{group} consists of a set $G$ together with an associative composition map $\circ \colon G \times G \to G$.  The composition map has an identity $1 \in G$ and composition with any element of $G$ is required to be invertible. A group $G$ has an \textbf{action} on a set $S$ if there is an action map $\cdot \colon G \times S \to S$ which is compatible with the composition law.  We say further that $\rho : G \mapsto GL(V)$ is a \textbf{$G$-representation} if the set $V$ is a vector space and each group element $g \in G$ is represented by a linear map (matrix) $\rho(g)$ that acts on $V$. Formally, a function $f \colon X \to Y$ may be described as respecting the symmetry coming from a group $G$ using the notion of equivariance. 

\begin{dfn}\label{dfn:equivariance}
Assume a group representation $\rho_{\text{in}}$ of $G$ acts on $X$ and $\rho_{\text{out}}$ acts on $Y$. We say a function $f$ is \textbf{$G$-equivariant} if 
\begin{equation}
    f( \rho_{\text{in}}(g)(x)) = \rho_{\text{out}}(g) f(x) \label{eq:strictequivariance}
\end{equation}
for all $x \in X$ and $g \in G$. The function $f$ is \textbf{$G$-invariant} if $f( \rho_{\text{in}}(g)(x)) = f(x)$ for all $x \in X$ and $g \in G$.  This is a special case of equivariance for the case $\rho_{\mathrm{out}}(g) = 1$. See Figure \ref{fig:equi} for an illustration of a rotation equivariant function. 
\end{dfn}

\subsection{Case Study: Equivariant Deep Dynamics Models}
\cite{Wang2020symmetry} exploited the symmetries of fluid dynamics to design equivariant networks. The Navier-Stokes equations are invariant under the following five different transformations. Individually, each of these types of transformations generates a group of symmetries in the system. 
\begin{itemize*}
    \item Space translation: \; $T_{\bm{c}}^{\mathrm{sp}}\bm{w}(\bm{x}, t) = \bm{w}(\bm{x-c}, t)$, \; $\bm{c} \in \mathbb{R}^2$,
    \item Time translation: \;$T_{\tau}^{\mathrm{time}}\bm{w}(\bm{x}, t) = \bm{w}(\bm{x}, t-\tau)$, \; $\tau \in \mathbb{R}$,
    \item Galilean transformation: \;$T_{\bm{c}}^{\mathrm{gal}}\bm{w}(\bm{x}, t) = \bm{w}(\bm{x}-\bm{c}t, t) + \bm{c}$, \; $\bm{c} \in \mathbb{R}^2$,
    \item Rotation/Reflection: \; $T_R^{\mathrm{rot}}\bm{w}(\bm{x}, t) = R\bm{w}(R^{-1}\bm{x}, t), \; R \in O(2)$,
    \item Scaling: \; $T_{\lambda}^{sc}\bm{w}(\bm{x},t) = \lambda\bm{w}(\lambda\bm{x}, \lambda^2t)$, \; $\lambda \in \mathbb{R}_{>0}$.
\end{itemize*}

Consider a system of 
differential operators $\mathcal{D}$ acting on $\hat{\mathcal{F}}_V$.  Denote the set of solutions $\mathrm{Sol}(\mathcal{D}) \subseteq \hat{\mathcal{F}}_V.$ We say $G$ is \textbf{a symmetry group of $\mathcal{D}$} if $G$ preserves $\mathrm{Sol}(\mathcal{D})$. That is, if $\varphi$ is a solution of $\mathcal{D}$, then for all $g \in G$, $g(\varphi)$ is also. 
In order to forecast the evolution of a system $\mathcal{D}$, we model the forward prediction function $f$. Let $\bm{w} \in \mathrm{Sol}(\mathcal{D})$.  The input to $f$ is a collection of $k$ snapshots at times $t - k,\ldots,t-1$ denoted $\bm{w}_{t-i} \in  \mathcal{F}_d$.  The prediction function $f\colon \mathcal{F}_d^k \to \mathcal{F}_d$ is defined $f(\bm{w}_{t-k},\ldots,\bm{w}_{t-1}) = \bm{w}_{t}$.  It predicts the solution at a time $t$ based on the solution in the past. 
Let $G$ be a symmetry group of $\mathcal{D}$.  Then for $g \in G$, $g(\bm{w})$ is also a solution of $\mathcal{D}$.  Thus  $f(g\bm{w}_{t-k},\ldots,g\bm{w}_{t-1}) = g\bm{w}_{t}$.  Consequently, $f$ is $G$-equivariant.

They tailored different methods for incorporating each symmetry into CNNs for spatiotemporal dynamics forecasting. CNNs are time translation-equivariant when used in an autoregressive manner. Convolutions are also naturally space translation equivariant. Scale equivariance in dynamics is unique as the physical law dictates the scaling of magnitude, space and time simultaneously. To achieve this, they replaced the standard convolution layers with group correlation layers over the group $G=(\mathbb{R}_{>0},\cdot)\ltimes(\mathbb{R}^2,+)$ of both scaling and translations. The $G$-correlation upgrades this operation by both translating \textit{and} scaling the kernel relative to the input, 

\begin{equation}\label{eqn:groupcorr}
 \bm{v}(\bm{p}, s, \mu) = \sum_{\lambda \in \mathbb{R}_{>0}, t \in \mathbb{R}, \bm{q} \in \mathbb{Z}^2 } \mu   \bm{w}(\bm{p} + \mu \bm{q}, \mu^2 t,    \lambda ) K(\bm{q},s,t,\lambda),
\end{equation}
where $s$ and $t$ denote the indices of output and input channels. They add an axis to the tensors corresponding to the scale factor $\mu$. In addition, the rotational symmetry was modeled using $\mathrm{SO}(2)$-equivariant convolutions and activations within the \texttt{E(2)-CNN} framework \citep{weiler2019e2cnn}.

To make CNNs equivariant to Galilean transformation, since they are already translation-equivariant, it is only necessary to make them equivariant to uniform motion transformation, which is adding a constant vector field to the vector field. This is part of Galilean invariance and relevant to all non-relativistic physics modeling. And the uniform motion equivariance is enforced by conjugating the model with shifted input distribution. Basically, for each sliding local block in each convolutional layer, they shift the mean of the input tensor to zero and shift the output back after convolution and activation function per sample. In other words,  if the input is $\bm{\mathcal{P}}_{b \times d_{in} \times s\times s} $ and the output is $\bm{\mathcal{Q}}_{b \times d_{out}} = \sigma(\bm{\mathcal{P}} \cdot K)$ for one sliding local block, where $b$ is batch size, $d$ is number of channels, $s$ is the kernel size, and $K$ is the kernel, then
\begin{align}
\bm{\mu}_i = \mathrm{Mean}_{jkl} \left(\bm{\mathcal{P}}_{ijkl}\right); \quad
\bm{\mathcal{P}}_{ijkl}\mapsto \bm{\mathcal{P}}_{ijkl} - \bm{\mu}_i; \quad
\bm{\mathcal{Q}}_{ij} \mapsto \bm{\mathcal{Q}}_{ij} + \bm{\mu}_i.
\end{align}
This will allow the convolution layer to be equivariant with respect to uniform motion. If the input is a vector field, this operation is applied to each element.

\begin{figure*}[htb!]
  \begin{minipage}[b]{0.241\textwidth}
   \includegraphics[width=\textwidth]{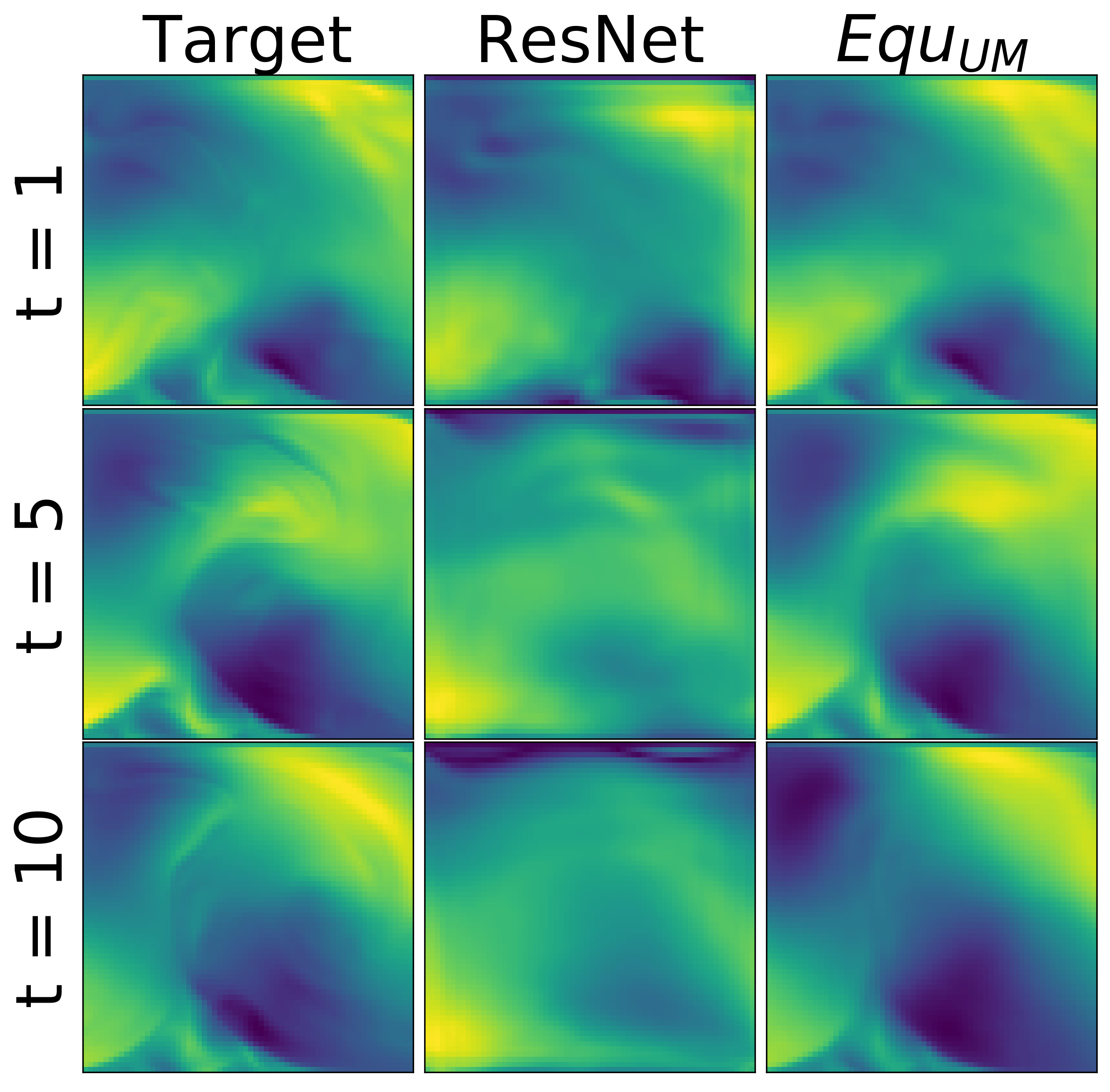}
  \end{minipage} \hfill
  \begin{minipage}[b]{0.241\textwidth}
     \includegraphics[width=\textwidth]{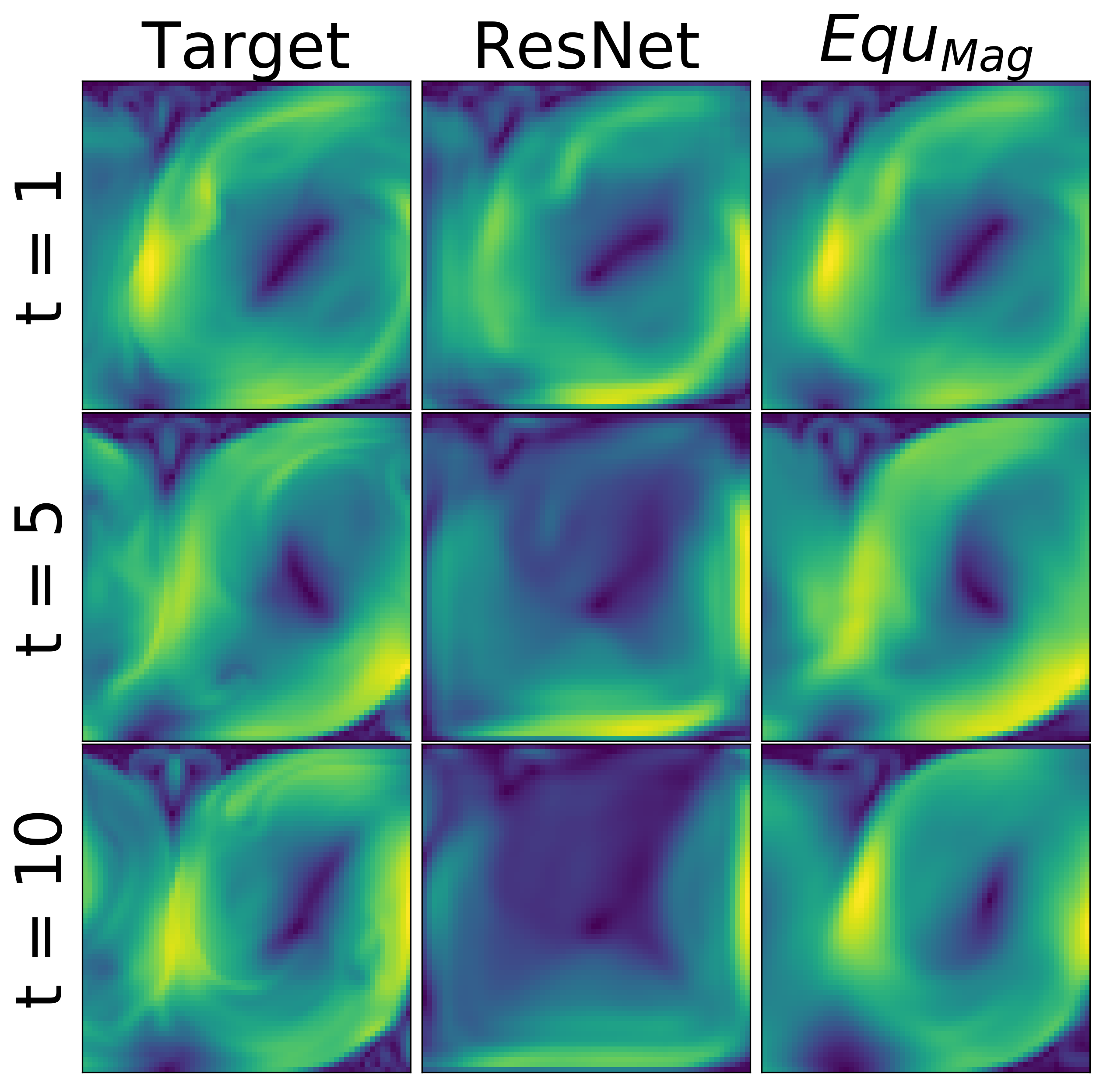}
  \end{minipage} \hfill
  \begin{minipage}[b]{0.241\textwidth}
\includegraphics[width=\textwidth]{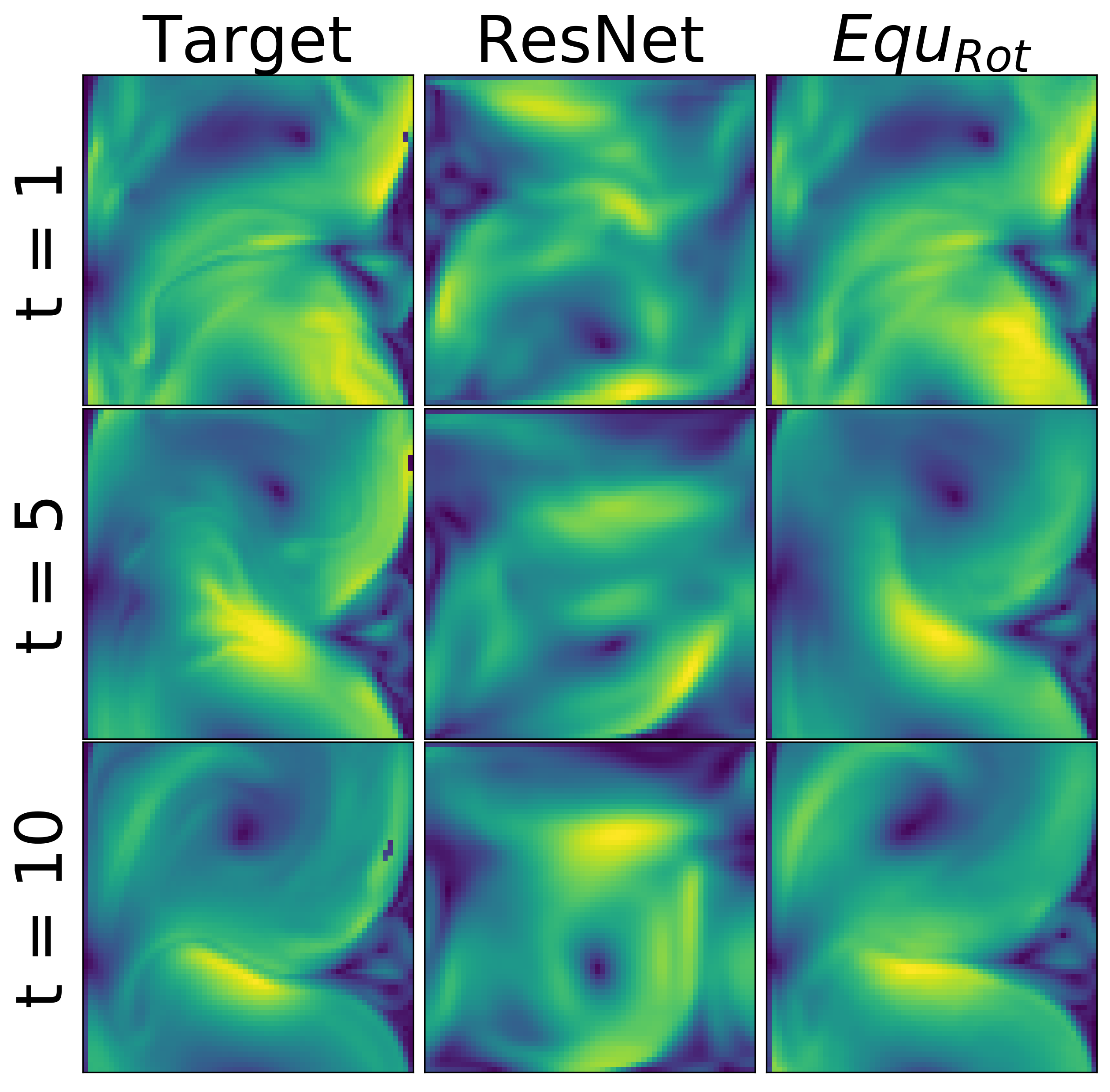}
  \end{minipage} \hfill
  \begin{minipage}[b]{0.241\textwidth}
\includegraphics[width=\textwidth]{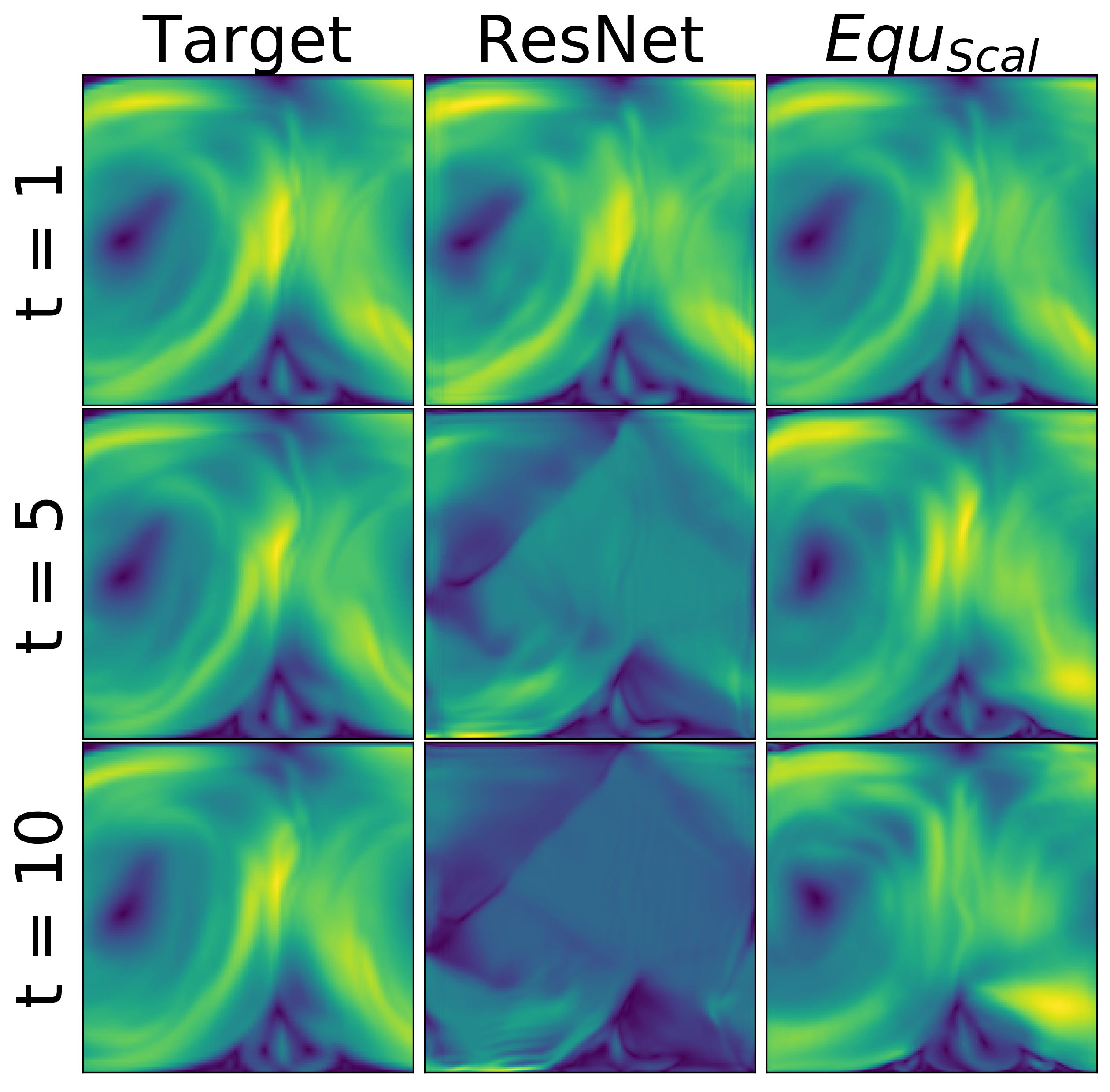}
  \end{minipage} 
    \begin{minipage}[b]{\textwidth}
\includegraphics[width=\textwidth]{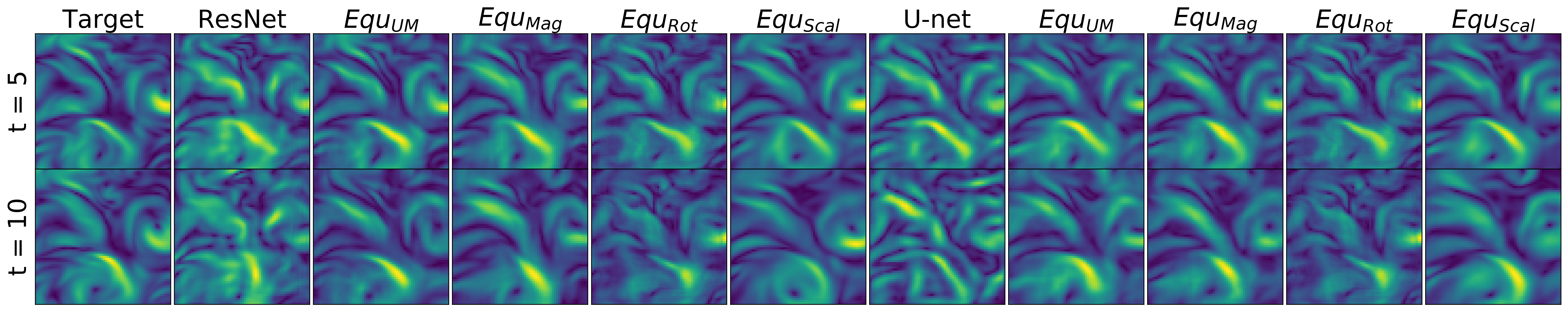}
  \end{minipage} 
\caption{Top: The ground truth and the predicted velocity norm fields $\|\bm{w}\|_2$ of RBC at time step $1$, $5$ and $10$ by the \texttt{ResNet} and four \texttt{Equ-ResNets} on four test samples applied with random uniform motion, magnitude, rotation, and scaling transformations respectively. The first column is the target, the second is \texttt{ResNet} predictions, and the third is predictions by \texttt{Equ-ResNets}. Bottom: The ground truth and predicted velocity norm fields of ocean currents by \texttt{ResNet} (\texttt{Unet}) and four \texttt{Equ-ResNets} (\texttt{Equ-Unets}) on the test set.}
\label{vel_u}
\end{figure*}

The DL models used are \textit{ResNet} and \textit{U-Net}, and their equivariant counterparts. Spatiotemporal prediction is done autoregressively. Standard RMSE and an RMSE computed on the energy spectra are used to measure performance. The models are tested on Rayleigh-B\'enard convection (RBC) and reanalysis ocean current velocity data. For RBC, the test sets have random transformations from the relevant symmetry groups applied to each sample. This mimics real-world data in which each sample has an unknown reference frame. For ocean data, tests are also performed on different time ranges and different domains from the training set, representing distributional shifts. Figure \ref{vel_u} shows the equivariant models perform significantly better than their non-equivariant counterparts on both simulated RBC data and real-world reanalysis ocean currents. They also show equivariant models also achieve much lower energy spectrum errors and enjoy favorable sample complexity compared with data augmentation. 

\subsection{Equivariant Convolutional Neural Networks}\label{fluid_sym}
However, real-world dynamical data rarely conform to strict mathematical symmetries, due to noise and missing values or symmetry-breaking features in the underlying dynamical system. \cite{wang2022approximately} further explored \textit{approximately} equivariant convolutional networks that are biased towards preserving symmetry but are not strictly constrained to do so. The key idea is relaxing the weight-sharing schemes by introducing additional trainable weights that can vary across group elements to break the strict equivariance constraints. The proposed approximate equivariant networks can always learn the correct amount of symmetry from the data, and thus consistently perform well on real-world turbulence data with no symmetry, approximate symmetry, and perfect symmetry.  When we incorporate prior knowledge into neural nets, we usually need to choose between strictly enforcing it in the design of the model or softly constraining it via regularizers. But this approach allows the model to decide whether and how to use prior knowledge (symmetry) based on the specific task. Moreover, \cite{Wang2021DyAd} built a meta-learning framework, DyAd, to forecast systems with different parameters. Specifically, it utilized an encoder capable of extracting the time-invariant and translation-invariant parts of a dynamical system and a prediction network to adapt and forecast giving the inferred system. Time invariance is achieved by using 3D convolution and time-shift invariant loss. On challenging turbulent flow prediction and real-world ocean temperature and currents forecasting tasks, this is the \textit{first} framework that can generalize and predict dynamics across a wide range of heterogeneous domains. 

Apart from incorporating symmetries into regular convolution, there has been a surge of interest in designing equivariant continuous convolution models. This is due to the fact that continuous convolution allows for convolutional operations to be performed on a continuous input domain. For instance, \cite{Schtt2017SchNetAC} proposed \textit{SchNet}, which is a continuous convolution framework that generalizes the CNN approach to continuous convolutions to model particles at arbitrary positions. Continuous convolution kernels are generated by dense neural networks that operate on the interatomic distances, which ensures rotational and translation invariance of the energy.  In a traffic forecasting application, \cite{walters2021trajectory} proposed a novel model, \textit{Equivariant Continuous COnvolution (ECCO)} that uses rotationally equivariant continuous convolutions to embed the symmetries of the system for improved trajectory prediction. The rotational equivariance is achieved by a weight-sharing scheme within kernels in polar coordinates. \textit{ECCO} achieves superior performance to baselines on two real-world trajectory prediction datasets, Argoverse and TrajNet++.

\subsection{Equivariant Graph Neural Networks} 
In addition to the equivariant convolution, numerous equivariant graph neural nets have also been developed, particularly for modeling atomic systems and molecular dynamics. This is due to the pervasive presence of symmetry in molecular physics, as evidenced by roto-translation equivariance in molecular conformations and coordinates. 

\cite{satorras2021n} designed E(n)-equivariant graph neural network for predicting molecular properties. It updates edge features with the Euclidean distance between nodes and updates the coordinates of particles with the weighted sum of relative differences of all neighbors. 
\cite{Shi2021Learning} proposed to use a score-based generative model for generating molecular conformation. The authors used equivariant graph neural networks to estimate the score function, which is the gradient fields of the log density of atomic coordinates because it is roto-translation equivariant. 
\cite{Anderson2019Cormorant} designed \textit{Cormorant}, a rotationally covariant neural network architecture for learning the behavior and properties of complex many-body physical systems. \textit{Cormorant} achieves promising results in learning molecular potential energy surfaces on the MD-17 dataset and learning the geometric, energetic, electronic, and thermodynamic properties of molecules on the GDB-9 dataset. 
\cite{simm2021symmetryaware} proposed a model for autoregressive generation of 3D molecular structures with reinforcement learning (RL). The method uses equivariant state representations for autoregressive generation, built largely from \textit{Cormorant}, and integrates such representations within an existing actor-critic RL generation framework.  
\cite{unke2021se} further designed a series SE(3)-equivariant operations and building blocks for DL architectures operating on geometric point cloud data, which was used to construct \textit{PhiSNet}, a novel architecture capable of accurately predicting wavefunctions and electronic densities.

Additionally, permutation invariance also exists in molecular dynamics. For instance, quantum mechanical energies are invariant if we exchange the labels of identical atoms. However, \cite{thiede2021autobahn} stated that enforcing equivariance to all permutations in graph neural nets can be very restrictive when modeling molecules. Thus, they proposed to decompose a graph into a collection of local graphs that are isomorphic to a pre-selected template graph so that the sub-graphs can always be canonicalized to template graphs before convolution is applied. By doing this, the graph neural nets can not only be much more expressive but also locally equivariant. \cite{villar2021scalars} proposed to build equivariant neural networks based on the idea that nonlinear $O(d)$-equivariant functions can be universally expressed in terms of a lightweight collection of scalars, which are simpler to build. They demonstrated the efficiency and scalability of their proposed approach to two classical physics problems, calculating the total mechanical energy of particles and the total electromagnetic force, that obeys all translation, rotation, reflection, and permutation symmetries. Moreover, since the design of equivariant layers is a difficult task, \cite{brandstetter2022lie} proposed a lie point symmetry data augmentation method for training graph neural PDE solvers and this method enables these neural solvers to preserve multiple symmetries.

\subsection{Symmetry Discovery} 
There has been also an emerging area that is symmetry discovery, the key idea of which is to find the weight-sharing patterns in neural networks that have been trained on data with symmetries. For instance, \cite{zhou2020meta} factorized the weight matrix in a fully connected layer into a symmetry (i.e. weight-sharing) matrix and a vector of filter parameters. The two parts are learned separately in the inner and outer loop training with the Model-Agnostic Meta-Learning algorithm (MAML) \citep{finn2017model}, which is an optimization-based meta-learning method so that the symmetry matrix can learn the weight-sharing pattern from the data. Furthermore, \cite{dehmamy2021automatic} proposed \textit{Lie Algebra Convolutional Network (L-conv)}, a novel architecture that can learn the Lie algebra basis and automatically discover symmetries from data. It can be considered as an infinitesimal version of group convolution. \cite{yang2023generative} further leveraged \textit{L-conv} to construct the \textit{LieGAN}, to automatically discover equivariances from a dataset using a paradigm akin to generative adversarial training. It represents symmetry as an interpretable Lie algebra basis and can discover various symmetries. Specifically, a generator learns a group of transformations applied to the data, which preserves the original distribution and fools the discriminator.

\subsection{Pros and Cons}
By designing a model that is intrinsically equivariant or invariant to input transformations, we can ensure that our model generalizes automatically across these transformations, making it resilient to distributional shifts. In contrast, data augmentation techniques cannot provide equivariance guarantees when the models are applied to new datasets. Empirically and theoretically, it has been shown that equivariant and invariant neural nets offer superior data and parameter efficiency compared to data augmentation techniques. Furthermore, incorporating symmetries enhances the physical consistency of neural nets because of Noether's Law.

However, incorporating too many symmetries may overly constrain the representation power of neural nets and slow down both training and inference. In addition, many real-world dynamics do not have perfect symmetries. A perfectly equivariant model that respects a given symmetry may have trouble learning partial or approximated symmetries in real-world data. Thus, an ideal model for real-world dynamics should be approximately equivariant and automatically learn the correct amount of symmetry in the data, such as the paper we discussed in Section \ref{fluid_sym}. There are a few other works that explore the same idea. For instance, \cite{finzi2021residual} proposed the soft equivariant layer by directly summing up a flexible layer with one that has strong equivariance inductive biases to model the soft equivariance.

\section{Discussion}\label{dis}
In this paper, we systematically review the recent progress in physics-guided DL  for learning dynamical systems. We discussed multiple ways to inject first-principle and physical constraints into DL  including (1) physics-informed loss regularizers (2) physics-guided design, (3) hybrid models, and (4) symmetry. By integrating physical principles, the DL models can achieve better physical consistency, higher accuracy, increased data efficiency, improved generalization, and greater interpretability. Despite the great promise and exciting progress in the field, physics-guided AI is still at its infant stage. Below we review the emerging challenges and opportunities of learning physical dynamics with deep learning for future studies.

\subsection{Improving Generalization}
Generalization is a central problem in machine learning. One current limitation of deep learning models for learning complex dynamics is their inability to understand the system solely from data and handle distributional shifts that naturally occur. Most deep learning models for dynamics modeling are trained to model a specific system and still struggle with generalization. For example, in turbulence modeling, deep learning models trained with fixed boundaries and initial conditions often fail to generalize to fluid flows with different characteristics. To overcome this limitation, one approach is to build physics-guided deep learning models, where the physics part plays a dominant role while the neural networks focus on learning the unknown process \cite{Rackauckas2020UniversalDE}. Another promising direction is meta-learning. For instance, \cite{Wang2021DyAd} proposed a model-based meta-learning method called \textit{DyAd} that can generalize across heterogeneous domains of fluid dynamics. However, this model can only generalize well on the dynamics with interpolated physical parameters and cannot extrapolate beyond the range of the physical parameters in the training set. Another idea is to transform data into a canonical distribution that neural networks can learn from and then restore the original data after predictions are made \cite{kim2022reversible}. Since neural networks struggle with multiple distributions, this approach aims to find a single distribution that can represent the dynamics effectively. A trustworthy and reliable model for learning physical dynamics should be able to extrapolate to systems with various parameters, external forces, or boundary conditions while maintaining high accuracy. Therefore, further research into generalizable physics-guided deep learning is crucial.

\subsection{Improving Robustness of Long-term Forecasting}
Long-term forecasting of physical dynamics is a challenging task as it is prone to error accumulation and instability to perturbations in the input, which significantly affect the accuracy of neural networks over a long forecasting horizon. To address these issues, several training techniques have been proposed in recent years. One such technique involves adding noise to the input, which makes the models less sensitive to perturbations \citep{brandstetter2022message}. It also suggests when making predictions in an autoregressive manner, the neural nets should be trained to make multiple steps of predictions in each autoregressive call instead of just one step.  Additionally, \cite{Rong2022Lyapunov} proposed a time-based Lyapunov regularizer to the loss function of deep forecasters to avoid training error propagation and improve the trained long-term prediction. Moreover, \citep{Pfaff2021LearningMS, osti_1834708} utilized online normalization that is normalizing the current training sample using a running mean and standard deviation, which also increases the time horizon that the model can predict. These models are trained on a large amount of simulation data. However, for real-world problems, obtaining real-world data such as experimental data of jet flow can be expensive, which presents a significant challenge for improving the robustness of predictions on limited training data. In such cases, developing robust prediction models that can generalize well on limited training data is of great importance.

\subsection{Learning Dynamics in Non-Euclidean Spaces}
Spatiotemporal phenomena, from global ocean currents to the spread of infectious diseases, are examples of dynamics in non-Euclidean spaces, which means they cannot be easily represented using traditional Euclidean geometry. To address this issue, the field of geometric deep learning \cite{bronstein2021geometric}  has emerged. Geometric deep learning aims to generalize neural network models to non-Euclidean domains such as graphs and manifolds. However, most of the existing work in this field has been limited to static graph data. Thus, learning dynamics in non-Euclidean Ssaces is a promising direction, and geometric concepts, such as rent notions of distance, curvature, and parallel transport, must be taken into account when designing models. For example, when modeling the ocean dynamics on the earth, which is a sphere, we need to encode the gauge equivariance \cite{cohen2019gauge} in the design of neural nets since there is no canonical coordinate system on a sphere. 

\subsection{Theoretical Analysis}
The majority of literature on learning dynamics with DL focuses on the methodological and practical aspects. Research into the theoretical analysis of generalization
is lacking. Current statistical learning theory is based on the typical assumption that training
and test data are identically and independently distributed (i.i.d.) samples from some unknown
distribution \cite{zhang2017understanding, neyshabur2017exploring}. However, this assumption does not hold for most dynamical systems, where observations at different times and locations may be highly correlated. \cite{kuznetsov2020discrepancy} provided the discrepancy-based generalization guarantees
for time series forecasting. On the basis of this,  \cite{wang2022data} took the first step to derive generalization bounds for equivariant models and data augmentation in the dynamics forecasting setting. The derived upper bounds are expressed in terms of measures of distributional shifts and group transformations, as well as the Rademacher complexity. But these bounds are sometimes not very informative since many of the inequalities used can be loose. However, to better understand the performance of DL on learning dynamics, we need to derive
generalization bounds expressed in terms of the characteristics of the dynamics, such as the order and Lyapunov exponents. Deriving lower generalization bounds are also necessary since they reveal the best performance scenarios. Theoretical studies can also inspire research into model
design and algorithm development for learning dynamics.

\subsection{Causal Inference in Dynamical Systems}
A fundamental pursuit in science is to identify causal relationships. In the context of dynamical systems, we may ask which variables directly or indirectly influence other variables through intermediates. While traditional approaches to the discovery of causation involve conducting controlled real experiments \citep{Pearl2009CausalII, Bollt2018IntroductionTF}, data-driven approaches have been proposed to identify causal relations from observational data in the past few decades \citep{Harradon2018CausalLA, Guo2020ASO}. However, most data-driven approaches do not directly address the challenge of learning causality with big data. Many questions remain open, such as using causality to improve deep learning models, disentangling complex and multiple treatments, and designing the environment to control the given dynamics. Additionally, we are also interested in understanding the system’s response under interventions. For example, when using deep learning to model climate dynamics, we need to make accurate predictions under different climate policies, such as carbon pricing policies and the development of clean energy, to enable better decisions by governments for controlling climate change.

\subsection{Search for Physical Laws}
Another promising direction is to seek physics laws with the help of DL. The search for fundamental laws of practical problems is the main theme of science. Once the governing equations of dynamical systems are found, they allow for accurate mathematical modeling, increased interpretability, and robust forecasting. However, current methods are limited to selecting from a large dictionary of possible mathematical terms \citep{rao2022discovering, Schmidt2009DistillingFN, Brunton2015Sparse, Lagergren2020LearningPD, Rudy2016pde}. The extremely large search space, limited high-quality experimental data, and overfitting issues have been critical concerns. Another line of work is to discover symmetry from the observed data instead of the entire dynamics with the help of DL \cite{dehmamy2021automatic, finn2017model}. But these works can only work well on synthetic data and discover known symmetries. Still, research on data-driven methods based on DL for discovering physics laws is quite preliminary. 


\subsection{Efficient Computation}
Given the rapid growth in high-performance computation, we need to improve automation and accelerate streamlining of highly compute-intensive workflows for science. We should focus on how to efficiently train, test, and deploy complex physics-guided DL models on large datasets and high-performance computing systems, such that these models can be quickly utilized to solve real-world scientific problems. To really revolutionize the field, these DL tools need to become more scalable and transferable and converge into a complete pipeline for the simulation and analysis of dynamical systems. A simple example is that we can integrate machine learning tools into the existing numerical simulation platforms so that we do not need to move data between systems every time and we can easily use either or both types of methods for analyzing data.

In conclusion, given the availability of abundant data and rapid growth in computation, we envision that the integration of physics and DL will play an increasingly essential role in advancing scientific discovery and addressing important dynamics modeling problems.

\section*{Acknowledgement}
This work was supported in part by U.S. Department Of Energy, Office of Science under grant DESC0022331, U. S. Army Research Office under Grant W911NF-20-1-0334, Facebook Data Science Award, Google Faculty Award, and NSF Grant \#2037745.

\clearpage
\bibliography{refs}
\end{document}